\documentclass[journal]{IEEEtran}
\usepackage{amsmath}
\usepackage{color}
\usepackage{multicol}
\usepackage{array}
 \usepackage[normalem]{ulem}
\usepackage{stfloats}
\usepackage{amsthm} 
\usepackage{multirow}
\usepackage{amssymb}
\usepackage{float}
\usepackage{nccmath} 
\usepackage{graphicx}

\usepackage{setspace}
\usepackage{booktabs}
\usepackage{mwe}
\usepackage{cite}
\usepackage{amsmath,amssymb,amsfonts}
\usepackage{graphicx}
\usepackage{textcomp}
\usepackage{xcolor}
\usepackage{multirow}
\usepackage{mathtools} 
\usepackage{physics}
\usepackage{bm}
\usepackage[utf8]{inputenc}
\usepackage[linesnumbered,ruled]{algorithm2e}
\usepackage{algorithmicx}
\usepackage{xspace}
\usepackage{subfigure}
\usepackage{hyperref}
\usepackage{tikz}

\usepackage{lipsum}

\makeatletter
\newsavebox{\@brx}
\newcommand{\llangle}[1][]{\savebox{\@brx}{\(\m@th{#1\langle}\)}%
  \mathopen{\copy\@brx\kern-0.5\wd\@brx\usebox{\@brx}}}
\newcommand{\rrangle}[1][]{\savebox{\@brx}{\(\m@th{#1\rangle}\)}%
  \mathclose{\copy\@brx\kern-0.5\wd\@brx\usebox{\@brx}}}
\makeatother

\newtheorem{assumption}{Assumption}
\newtheorem{lemma}{Lemma}
\newtheorem{theorem}{Theorem}

\newtheorem{proposition}{Proposition}
\usepackage{amsmath,amsfonts,bm}

\def\eqref#1{eqn.~(\ref{#1})}
\def\ceil#1{\lceil #1 \rceil}

\def\1{\bm{1}}

\DeclareMathAlphabet{\mathsfit}{\encodingdefault}{\sfdefault}{m}{sl}
\SetMathAlphabet{\mathsfit}{bold}{\encodingdefault}{\sfdefault}{bx}{n}

\def\tt{{\bf t}}

\def\0{{\bf 0}}
\def\1{{\bf 1}}

\def\EB{{\mathbb E}}

\begin{document}
\title{SlimFL: Federated Learning with Superposition Coding over Slimmable Neural Networks}

\author{Won Joon Yun, Yunseok Kwak, Hankyul Baek, Soyi Jung,~\IEEEmembership{Member,~IEEE,} Mingyue Ji,~\IEEEmembership{Member,~IEEE,} \\ Mehdi Bennis,~\IEEEmembership{Fellow,~IEEE,} Jihong Park,~\IEEEmembership{Senior Member,~IEEE,} and Joongheon Kim,~\IEEEmembership{Senior Member,~IEEE}
\thanks{Preliminary version of this paper was accepted to IEEE Conference on Computer Communications (INFOCOM), May 2022~\cite{infocom2022baek}.}
\thanks{This research is supported by the National Research Foundation of Korea (NRF-Korea, 2022R1A2C2004869) and the Institute of Information \& communications Technology Planning \& Evaluation (IITP) grant funded by the Korea government (MSIT) (No. 2021-0-00467, Intelligent 6G Wireless Access System). \typeout{\textit{(Won Joon Yun, Yunseok Kwak, and Hankyul Baek contributed equally to this work (first authors).)}} \textit{(Corresponding authors: Soyi Jung, Jihong Park, Joongheon Kim)}}
\thanks{W. J. Yun, Y. Kwak, H. Baek, and J. Kim are with the School of Electrical Engineering, Korea University, Seoul 02841, Republic of Korea, e-mails: \{ywjoon95,rhkrdbstjr0,67back,joongheon\}@korea.ac.kr.}
\thanks{S. Jung is with the Department of Electrical and Computer Engineering, Ajou University, Suwon 16499, Republic of Korea, e-mail: sjung@ajou.ac.kr.}
\thanks{M. Ji is with the Department of Electrical and Computer Engineering, The University of Utah, Salt Lake City, UT 84112, USA, e-mail: mingyue.ji@utah.edu.}
\thanks{M. Bennis is with the Centre for Wireless Communications, University of Oulu, Oulu 90014, Finland, e-mail: mehdi.bennis@oulu.fi.} % .
\thanks{J. Park is with the School of Information Technology, Deakin University, Geelong, VIC 3220, Australia, e-mail: jihong.park@deakin.edu.au.}
}
\maketitle

\begin{abstract}
Federated learning (FL) is a key enabler for efficient communication and computing, leveraging devices' distributed computing capabilities.
However, applying FL in practice is challenging due to the local devices' heterogeneous energy, wireless channel conditions, and non-independently and identically distributed (non-IID) data distributions. 
To cope with these issues, this paper proposes a novel learning framework by integrating FL and width-adjustable slimmable neural networks (SNN). 
Integrating FL with SNNs is challenging due to time-varying channel conditions and data distributions.
In addition, existing multi-width SNN training algorithms are sensitive to the data distributions across devices, which makes SNN ill-suited for FL. Motivated by this, we propose a communication and energy-efficient SNN-based FL (named \textit{SlimFL}) that jointly utilizes \textit{superposition coding (SC)} for global model aggregation and \textit{superposition training (ST)} for updating local models. By applying SC, SlimFL exchanges the superposition of multiple-width configurations decoded as many times as possible for a given communication throughput. Leveraging ST, SlimFL aligns the forward propagation of different width configurations while avoiding inter-width interference during backpropagation. We formally prove the convergence of SlimFL. The result reveals that SlimFL is not only communication-efficient but also deals with non-IID data distributions and poor channel conditions, which is also corroborated by data-intensive simulations.
\end{abstract}

\IEEEpeerreviewmaketitle
\section{Introduction}\label{sec:1}
\subsection{Background and Motivation}
\IEEEPARstart{R}{ecent} advances in machine learning (ML), and hardware technologies have pushed deep learning down from cloud servers to edge devices such as phones, cars, and the Internet of things (IoT) devices~\cite{savazzi2021opportunities,isj21dao}. These edge devices collectively constitute a source of ever-growing big data, so are indispensable for training high-quality machine learning models. Meanwhile, each edge device stores only a tiny fraction of the big data, which is often privacy-sensitive (e.g., navigation history, e-Health wearable records, and surveillance camera photos). To exploit these opportunities and address the challenges induced by edge devices, federated learning (FL) is a  promising solution that allows edge devices to train a global model by exchanging locally trained models instead of raw data~\cite{Brendan17,infocom1,tonfl1,9067847}. 

At its core, FL operations rest on repeatedly constructing a global model averaged over the local models, which is downloaded by each device to replace its local model with the new global model. By design, FL necessitates the use of a single global model for all devices. This requirement restricts the scalability and accuracy of FL, particularly when scaling to the sheer number of edge devices, which have non-identical memory resources and energy budgets not to be mentioned as different communication channel conditions. Therefore, the use of a single large model may be suited for only a few devices, whereas the use of a tiny model should be accompanied by compromising accuracy. 

In light of the aforementioned issues in communication-energy efficiencies and scalability, the recently proposed slimmable neural network (SNN) architectures have great potential in that an SNN can adjust its model width in accordance with its available energy budget or task difficulty. Inspired from this, in this article, we propose an \textit{SNN-based FL algorithm} that makes use of \textit{superposition coding (SC)} and \textit{successive decoding (SD)}, coined \textit{slimmable FL (SlimFL)}. 
To illustrate the effectiveness of SlimFL, consider an SNN with two width configurations, as seen in Fig.~\ref{fig:abstract}. Each device uploads its local updates during the uplink to the server after jointly encoding the left-half (LH) and right-half (RH) of its local SNN model and assigning distinct transmission power levels, i.e., SC~\cite{Cover:TIT72}. The server then makes an attempt to decode the LH. If the LH is successfully decoded, the server attempts to decode the RH consecutively, i.e., SD, also known as successive interference cancellation (SIC). As a result, when the device-server channel throughput is poor, the server decodes only the LH of the uploaded model, yielding a model with a \textit{half-width (0.5x) width}. When the channel has a high throughput, the server may decode both LH and RH and combine them to produce the \textit{full-width model (1.0x)}. As a result, the server creates a global model superimposing the decoded 0.5x and 1.0x local models, which each device downloads. The device substitutes the downloaded global model for its local model and repeats the preceding step until convergence. In addition, the same technique may be extended to the server-to-device downlink.

SlimFL's efficacy is contingent upon achieving synergy between several width configurations, namely 0.5x and 1.0x models, which is a difficult task. The existing FL algorithm does not focus on synergetic of 0.5x and 1.0x. In addition, since the local model has various width configurations, training them may conflict with one another.
To the best of our knowledge, FL with different width configurations have not been proposed, yet. Existing SNN architectures and training techniques are designed for standalone learning and so are unsuitable for SlimFL, especially when data distributions are not independent and identical (non-IID). To address these issues in SlimFL, we provide a new SNN architecture and training method called \textit{superposition training (ST)}, as well as a study of SlimFL's convergence and efficiency.

\subsection{Contributions}
\begin{figure*}[t!]
\centering
\includegraphics[width=0.9\textwidth]{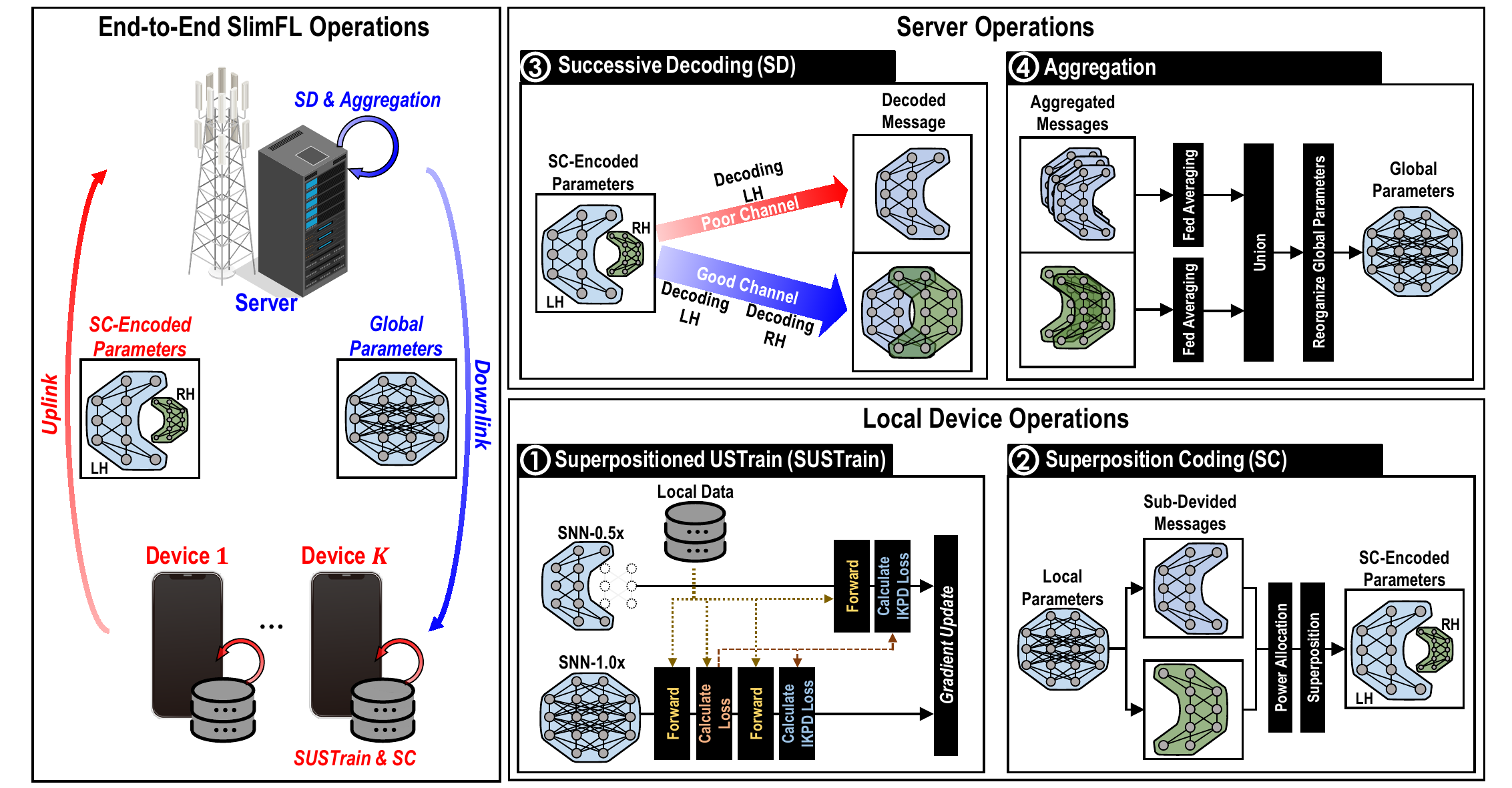}
\caption{A schematic illustration: SlimFL consists of four processes. On the local side, the local SNN is trained with SUSTrain. After training, the local SNN is coded with SC. The local devices transmit SC-encoded local parameters to the server. On the server side, the server successively decodes SC-encoded parameters. The server aggregates the decoded parameters and reconstructs the global parameters.}
    \label{fig:abstract}
    \vspace{-5mm}
\end{figure*}

The major contributions of this paper are summarized as follows.
\begin{itemize}
\item We first propose an FL framework for SNNs, SlimFL (see Fig.~\ref{fig:abstract} and \textbf{Algorithm~\ref{alg:SlimFL}}), which uses SC for enhancing communication efficiency under time-varying wireless channels with limited bandwidth.
\item We develop a local SNN training method for SlimFL, ST (see \textbf{Algorithm~\ref{alg:sustrain}}), which minimizes excessive inter-width interference and achieves fast convergence with good accuracy while being agnostic to data distribution.
\item We prove the convergence of SlimFL (see \textbf{Theorem~\ref{convth2}}). Numerical results show the benefits of SlimFL in terms of channel quality and data distributions and characterize the optimum transmit power allocation for SC (see \textbf{Proposition~1}), and the optimal ST ratio allocation (see \textbf{Proposition~2}).
\item We verify our analysis via simulations, indicating that SlimFL achieves higher accuracy and lower communication costs than vanilla FL (i.e., FedAvg) under poor channel conditions and non-IID data distributions.
\end{itemize}

In our previous work~\cite{infocom2022baek}, we only described the brief idea of SlimFL and provided the proof sketches of convergence. In this work, we delineate the detailed operations and motivation of SlimFL, while elaborating on the full derivation steps of the convergence proofs. To further advocate the feasibility of SlimFL under various scenarios, we numerically study the impacts of different channel models, datasets, and local training capabilities in Sec.~\ref{sec:6}-D.

\subsection{Organization}
The rest of this paper is organized as follows.
Sec.~\ref{sec:2} describes previous research results of FL and background knowledge of superposition coding and successive decoding. Sec.~\ref{sec:3} presents the proposed SNN model and its training method (i.e., SUSTrain), and Sec.~\ref{sec:4} designs SlimFL using superposition coding, and successive decoding, Sec.~\ref{sec:5} shows the convergence analysis on SlimFL. Sec.~\ref{sec:6} presents the simulation-based performance evaluation and its result regarding SlimFL. 
Finally, Sec.~\ref{sec:7} concludes this paper. 

The notations used in this paper are listed in Table~\ref{tab:notation-convergence}.

\begin{table}[t!]
    \caption{List of Notations}
    \label{tab:notation-convergence}
    \centering
    \footnotesize
    \begin{tabular}{c|l}
        \toprule[1pt]
        \textbf{Symbol} & \textbf{Description}\\\midrule[1pt]
        ${K}$ & The number of devices\\
        ${T}$ & Total iteration step\\
        ${S}$ & The number of width configurations in SNN\\
        $~\theta^{G}$ & Parameter of global model \\
        $~\theta^{k}$ & Parameter of $k$-th local model \\
        \midrule
        $~\Xi_{{i}}$ & Binary mask to extract model parameter of $i$-th \typeout{small }model \\
        $\Xi$ & Binary mask to extract model parameter of LH segment \\
        $~~~~\Xi^{-1}$ & Binary mask to extract model parameter of RH segment \\
        $\mathsf{H}$ & Set of successfully decoded LH segment \\
        $\mathsf{F}$ & Set of successfully decoded RH segment \\
        $n_\mathsf{L}$ & The number of successfully decoded LH segments \\
        $n_\mathsf{R}$ & The number of successfully decoded RH segments \\
        $p_i$ & Decoding success probability of $i$-th message \\
        \midrule     
        $\bm{Z}$ & Entire dataset \\
        $\zeta^k_t$ & Local data sampled from $k$-th device at $t$ \\ 
        $w_{i}$ &  Ratio of superpositioned training for updating $i$-th \typeout{small }model \\ 
        $\lambda$ & Power allocation ratio \\ 
        $\eta_t$ & Learning rate of $t$\\ 
        \midrule   
        $\odot$ & Element-wise multiplication \\
        $M(\theta^k_t,\zeta^k_t)$ & Logits from feed-forwarding $\zeta^k_t$ to $k$-th SNN \\
        $y(\zeta^k_t)$ & Ground truth of $\zeta^k_t$\\
        \midrule   
        $\gamma$ & Signal-to-interference-plus-noise ratio (SINR) \\
        $~~\sigma^2$ & Noise power \\
        $d$ & Distance between the local device and the server \\
        $\beta$ & Pathloss exponent \\
        $P$ & Transmit power \\
        $~P_I$ & Interference power \\
        $u$ & Code rate \\
        $\chi$ & Small-scale fading \\
        $R$ & Received throughput \\
        \bottomrule[1pt]
     \end{tabular}
\end{table}

\section{Related Work}\label{sec:2}

\subsection{Energy-Constrained Federated Learning} 
In FL, each edge device trains its model with its own data. After training, all parameters of each local model are transmitted to the server (e.g., cloud, fog, or edge). The server reconstructs the global model by aggregating the local model parameters.  FL is privacy-preserving because the device does not transmit its own data to the server. FL has advantages regarding communication/computing costs as well as privacy-preserving.
For example, \textit{Federated Averaging} (FedAvg) reduces 10-100x of communication costs compared to the existing distributed machine learning algorithm~\cite{GoogleBlog,Brendan17}. In addition, FL with local batch normalization (FedBN~\cite{FedBN}), FL with a generalization and re-parametrization (FedProx~\cite{FedProx}), or adaptative aggregation (FedOpt~\cite{FedOpt}) improves the speed of convergence of the FL regime, which means reducing more communication/computing costs through training algorithms. 
However, the reduced communication/computing cost is non-negligible, because the communication cost is proportional to the model size. To cope with this problem, one approach is to optimize resource allocation~\cite{yang2020energy,chen2020joint}. The other approach is split learning~\cite{koda2020communication}, which reduces communication costs by transmitting smashed data rather than model parameters.
The other option is to investigate deep learning technique, \textit{e.g.,} model pruning to satisfy varying on-device energy and memory limits \cite{Han:16} or to distill the knowledge of a big trained model into a small empty model through knowledge distillation (KD) \cite{HintonKD:14}, but this requires extra training operations. Alternatively, one may alter the width and/or depth of a trained model to match the resource needs. After training, depth-controlled neural networks \cite{IJCNN2019_DepthControllable} and adaptive neural networks \cite{AAAI2019_Anytime} can modify their depths, while SNNs alter their widths   \cite{ICCV2019_USlimmable}.
According to studies \cite{yu2018slimmable,ICCV2019_USlimmable,li2021dynamic}, many SNN architectures and their algorithms are proposed. In this paper, we leverage width-controllable SNNs, and develop its FL version, SlimFL. Such an extension is non-trivial and entails several design issues, such as local SNN training algorithms, aggregating segment prioritization\typeout{ (e.g., more aggregating the same LH/RH segments vs. balancing the LH and RH aggregations), which will be discussed in Sec~\ref{sec:4}}.

\subsection{Superposition Coding \& Successive Decoding}
Non-orthogonal multiple access (NOMA) often outperforms orthogonal multiple access (OMA) in terms of sum rate and outage probability~\cite{higuchi2015non,choi2015minimum,Jinho16}. While uplink NOMA leverages distance and/or power differences to support more devices~\cite{7842433,iotj21tuong}, downlink NOMA commonly focuses on simultaneously broadcasting multiple signals with different priorities, such that the higher priority signal is more likely to be successfully decoded. Namely, the base station in downlink NOMA superimposes distinct data signals over the same radio block (\textit{i.e.}, SC) by assigning different transmit power levels~\cite{Cover:TIT72}. 
The mobile user decodes the SC-encoded signal through successive interference cancellation that decodes the strongest signal first, followed by decoding the next strongest signal after removing the decoded signal~(\emph{i.e.}, SD) \cite{TseBook:FundamaentalsWC:2005}. 

Inspired from downlink NOMA, in this paper we apply SC and SD for the uplink operations of SlimFL. Precisely, SlimFL prioritizes an SNN's LH to obtain the 0.5x model across inferior channels. SlimFL is capable of decoding the RH of the SNN only under favorable the channel circumstances, yielding the 1.0x model by merging both LH and RH. As a result, SlimFL provides consistent convergence even in the presence of weak channels.

\section{SNN Model Architecture and Training}\label{sec:3}
Currently, available SNN architectures and training algorithms are optimized for standalone learning \cite{ICCV2019_USlimmable}. This section presents SNN architecture for SlimFL, and its local training.

\subsection{Operations of Slimmable Neural Network}\label{sec:3-1}
Lightweight deep learning techniques have been studied widely. 
Thus many complicated models are distributed to mobile devices via quantization, e.g., TensorRT~\cite{gholami2021survey}.
Nevertheless, training models in mobile devices still requires computational cost as same as the existing methods. 
We utilize SNNs as both devices and server models. The key idea of SNN is to reduce floating point operations per second (FLOPS) by activating the selected weights. We compare the existing NN operation to SNN's.
Suppose that input and output pairs are denoted as $\mathbf{w}$ and $\mathbf{v}$, respectively. The linear operation is written as follows:
\begin{equation}
    \mathbf{v} = \mathbf{W} \cdot \mathbf{w} + \mathbf{b},
\end{equation} where $\mathbf{W}$, $\mathbf{b}$ denote $|\mathbf{v}|\times|\mathbf{w}|$ sized matrix and $|\mathbf{v}|$ sized vector, respectively. 
 Compared to linear operation, the slimmable linear operation utilizes the shrink ratio is denoted as $r \in \mathbb{R}[0,1]$ to reduce FLOPS. In the linear operation, the output dimension is reduced by zero-masking $\mathbf{W}$ and $\mathbf{b}$. The following masked weights and biases are obtained as follows:
 \begin{equation}
 \mathbf{W}_{\text{masked},r} =  \mathsf{Mask}(\mathbf{W}, r),~~ \mathbf{b}_{\text{masked},r} =  \mathsf{Mask}(\mathbf{b}, r),
 \end{equation} where $\mathsf{Mask}(\mathbf{W}, r)$ and $\mathsf{Mask}(\mathbf{b}, r)$ reduces the size by tensor product into $\ceil{|\mathbf{v}|\cdot r}\times|\mathbf{w}|$ and $\ceil{|\mathbf{v}|\cdot r}$, respectively. 
 The mobile device can reduce the computational complexity in both training and interference phases up to $\mathcal{O}({r^2})$.

Hereafter, we consider two width configurations with the ratio $r=0.5$ and $r=1.0$ (i.e., 0.5x or 1.0x).
At the $t$-th iteration, the SNN model has the parameters $\theta_t^k$ with two width configurations: 0.5x width configuration $\theta_t^k \odot \Xi_1$ and 1.0x width configuration $\theta_t^k \odot \Xi_2$ ($= \theta_t^k$), where $\odot$ is the element-wise product and $\Xi_i$ represents a binary mask for extracting the parameters of $i$-th width configuration.

\subsection{Superposition SNN Training}\label{superposition}

\begin{algorithm}[ht]\label{alg:slim}
\small
\caption{SlimTrain~\cite{yu2018slimmable}} 
    {Define \textit{switchable width list} for slimmable network \(M\), for example, {\([0.25, 0.5, 0.75, 1.0]\times\)}.}\\
    {Initialize shared convolutions and fully-connected layers for slimmable network \(M\).}\\
    {Initialize independent batch normalization parameters for each \textit{width} in \textit{switchable width list}.}\\
    \For {$i = 1, ..., n_{iters}$}
        {Get next mini-batch of data \(x\) and label \(y\).\\
        Clear gradients of parameters, \(optimizer.zero\_grad()\).\\
        \For {\textit{width} in \textit{switchable width list}}
            {
            Switch the batch normalization parameters of current width on network \(M\).\\
            Execute sub-network $M'$ at current width, \(\hat{y} = M'(x)\).\\
            Compute loss, \(loss = criterion(\hat{y}, y)\).\\
            Compute gradients, \(loss.backward()\).
            }
        Update parameters, \(optimizer.step()\).
        }
\end{algorithm}

\begin{algorithm}[ht]\label{alg:usslim} 
\small
    Define \textit{width range}, for example, { $[0.25,0.5,0.75,1.0]$x}.\\
    Define \textit{n} as the number of sampled widths per training iteration, for example, $n=4$.\\
    Initialize training settings of shared network \(M\).\\
    \For {$(t = 1, ..., T_{iters})$}
    {
        Get the next mini-batch of data $x$ and label $y$).\\
        Clear gradients, $optimizer.zero\_grad()$.\\
        Execute full-network, $y' = M(x)$.\\
        Compute loss, $loss = criterion(y', y)$.\\
        Accumulate gradients, $loss.backward()$.\\
        Stop gradients of $y'$ as label, $y' = y'.detach()$.\\ 
        Add smallest width to \textit{width samples}.\\
        \For {\textit{width} in \textit{width samples}}{
            Execute sub-network at \textit{width}, $\hat{y} = M'(x)$.\\
            Compute loss, $loss = criterion(\hat{y}, y')$.\\
            Accumulate gradients, $loss.backward()$.
            }
        Update parameters, \(optimizer.step()\).
    }
\caption{USTrain~\cite{ICCV2019_USlimmable}}
\end{algorithm}

\begin{algorithm}[t!]\label{alg:sustrain}
\small
Initialize train parameter $\Theta =\{\theta^1,\cdots,\theta^k,\cdots,\theta^K,\theta^G\}$,\\
Initialize local dataset $\bm{Z}=\{Z_1,\cdots,Z_k,\cdots,Z_K\}$ with Dirichlet distribution\\
Initialize learning rate $\eta_t\leftarrow \eta_0$\\
Further Constraints: $\hat{F}^1(\cdot)=F^1(\cdot)$ \Comment{Discuss in Sec.~\ref{sec:5}}\\
\For{$t=1,\cdots,T$}
    {
    \For{$k=1,\cdots,K$}
        {
        Initialize gradients of the model optimizer as $0$.\\
        Sample batch $\zeta^k_t$ from $Z_k$.\\
        Execute full-network  $M(\theta_t^k,\zeta_t^k)$.\\
        Compute loss,         $loss \leftarrow F^k(\theta_t^k,\zeta_t^k)$.\\
        Accumulate gradients, $loss.backward()$.\\
        Execute full-network $M(\theta_t^k,\zeta_t^k)$.\\
        Stop gradients of  $M(\theta_t^k,\zeta_t^k)$ as label.\\
        \For{$i=1,\cdots,S-1$}
            {
            Execute and calculate loss $\hat{F}^k(\theta^k_t \odot \Xi_i, \zeta^k_t)$\\
            $loss \leftarrow loss  + w_i\hat{F}^k(\theta^k_t \odot \Xi_i, \zeta^k_t)$.
            }
            Calculate gradient of $loss$.\\ 
            Update model parameters.  \Comment{Eq.~(\ref{eq:SUSTrain})}
        }
    }
\caption{Superposition Training (SUSTrain)}
\end{algorithm}
Training a multi-width SNN is arduous. The SNN backpropagation (BP) is warped due to gradient interference of multi-width. For example, the BP of 0.5x width configurations interferences the BP of 1.0x's and vice versa.
This inter-width interference hinders not just inference accuracy but also training convergence. SlimTrain (see \textbf{Algorithm~\ref{alg:slim}}), the first SNN training algorithm introduced in \cite{yu2018slimmable}, partially ameliorates such inter-width interference by training alternative width configurations in descending order of size.
While adhering to the sample principle, the authors of \cite{ICCV2019_USlimmable} have proposed a state-of-the-art SNN training technique called universal SNN (USTrain).

USTrain proposes the sandwich rule with the inplace knowledge distillation (IPKD), which are two notable techniques for training an SNN. To describe two notable techniques, the $1.0$x model becomes a teacher guiding its sub-width models via knowledge distillation. By nature, it gives more benefits under a larger SNN (i.e., a better teacher) that has more sub-width configurations (i.e., more students).
 The IPKD encourages each sub-width (i.e., student) to provide a softmax output (i.e., logit) to that of the full-width (i.e., teacher), so that their overlapping BP gradients become less dissimilar, hence minimizing the inter-width interference. At its core, the inplace distillation and the sandwich rule are effective under the case where an SNN can be divided into more than two segments. The SNN architecture considered in this paper consists of only the LH and RH segments, making the USTrain unfit for our case. Moreover, during updating the local parameters of slimmable model, some of the parameters composing small model needs partial gradient calculation of big model.
 
 While effective in standalone learning, in SlimFL with wireless connectivity, not all multi-width configurations are exchanged due to insufficient communication throughput. In other words, they exchanged width configurations are aggregated across devices, diluting the effectiveness of BN.
In our experiments, we even observed training convergence failures due to BN. Furthermore, managing multiple BN layers not only consumes additional memory costs but also entails a high computing overhead.

To resolve this problem, we propose \textit{superpositioned USTrain (SUSTrain)} algorithm. SUSTrain consists of two processes. First, all the forward propagation losses (FP) are holden. Then, all the width configurations are concurrently updated with the superpositioned gradients. With SUSTrain, a sub-width configuration (i.e., student) is trained using IPKD without the logit mismatch with its full-width configuration's logit (i.e., teacher's logit), while the full-width configuration is simultaneously trained using the ground truth. 

In this paper, we consider that all devices have SNN with two-width configurations (i.e., one teacher and one student). Hereafter, we generalize the local SNN update rule for the device~$k$ described as follows:
\begin{multline}\label{eq:SUSTrain}
    \hspace{-5pt}\theta^k_{t+1} \!=\! \theta_t^k  \!-\! \eta_t \big[ w_1 \nabla \hat{F}^k (\theta_t^k\odot\Xi_1,\zeta^k_t) \!+\! w_2\nabla {F}^k (\theta^k_t \odot \Xi_2, \zeta^k_t)\big],
\end{multline}
where $w_1 \!+\! w_2 \!=\! 1$ and $w_1,\! w_2\!>\!0$. The operator $\odot$ implies element-wise multiplication for extracting the parameters which is allocated to the SNN width configuration. The term $\eta_t>0$ is a learning rate, and $\zeta_t^k$ implies a stochastic input realization. The function $F^k(\theta_t^k\odot \Xi_i,\zeta_t^k)$ is the cross-entropy between the ground truth $y(\zeta_t^k)$, whereas the IPKD function $\hat{F}^k(\theta_t^k\odot \Xi_i,\zeta_t^k)$ is the cross-entropy between the logit $M(\theta_t^k,\zeta_t^k)$ of the full-width configuration and the logit $M(\theta_t^k\odot \Xi_i,\zeta_t^k)$ of the $i$-th width configuration. The details of SUSTrain are in \textbf{Algorithm~\ref{alg:sustrain}}.

\section{Global Model Aggregation with Superposition Coding \& Successive Decoding}\label{sec:4}
\subsection{Superposition Coding \& Successive Decoding}
\subsubsection{Wireless Systems}
Consider a single base station equipped with a server. The server is associated with $K$ mobile devices at equal distance $d$, e.g., uniformly distributed around a circle centered at the server with a radius of $d$. Unless otherwise specified, hereafter we focus only on uplink communications. In the uplink, each device aims to upload $S$ messages corresponding to $S$ width configurations. For ease of explanation, we consider frequency division duplex (FDD) and orthogonal frequency division multiplexing (OFDM), which can be replaced with other duplexing and multiplexing schemes such as time division duplex (TDD) and time division multiplexing (TDM) \cite{Molisch} with minor modifications. For simplicity without loss of generality, we consider that the bandwidth is orthogonally and equally allocated to each device, thereby ignoring inter-user interference. Notwithstanding, we still consider the interference due to SC and SD operations. For mathematical amenability, we assume that the interference is treated as noise, and use the Shannon's capacity formula with the Gaussian codebook, i.e., the use of optimal source and channel coding, which are widely used assumptions for analysis \cite{TseBook:FundamaentalsWC:2005}.

\subsubsection{Superposition Coding (SC)}
Recall that each mobile device $k$ stores a local model having $S$ width configurations. When $S=2$, the model is equally split as the LH and the RH segments, and the half-width model (0.5x) is the LH
while the entire model (1.0x) is obtained by combining the LH and the RH. Generalizing this, we can treat $S$ width configurations as $S$ messages, and concurrently transmit them via SC \cite{Cover:TIT72}. To this end, consider that $P_i$ transmit power is allocated to the $i$-th message out of the total power budget $P$, i.e.,  $P = \sum\limits^{S}_{i=1}\nolimits P_{i}, \forall i\in[1,S]$. 
 Consider a mobile device $k$ transmitting a signal $\mathbf{x}_k$ to the server over the same radio resource block, which is as:
\begin{equation}
    \mathbf{x}_k = \sum^S_{i=1}\nolimits\mathbf{s}_{k,i},
\end{equation}
where $\mathbf{s}_{k,i} \in \mathcal{S}_k$ denotes the symbol $i$ of mobile device $k$, and $\mathcal{S}_k$ is the Gaussian codebook for device $k$, with $\mathbb{E}[\mathbf{s}_{k,i}] = 0$ and $\mathbb{E}[|\mathbf{s}_{k,i}|^2] = P_i$, $\forall i \in [1,S], \forall k \in [1,K]$.

\subsubsection{Decoding Success Probability and Successive Decoding~(SD)} 
In SD, the receiver first decodes the strongest signal. Then, it sequentially decodes the next strongest signal, after cancelling out the decoded signal while treating the rest as interference, i.e., successive interference cancellation~\cite{Jinho:TCOM17}. For simplicity, assume that $P_i > P_{i'}$ for all $i'>i$. Following SD, for the device $k$, the server first decodes $\mathbf{s}_{k,1}$ while treating the rest $P_{k,1}^I$ as interference, followed by decoding $\mathbf{s}_{k,2}$ in the presence of the interference $P_{k,2}^I \leq P_{k,1}^I$. Each decoding becomes successful when the throughput $R_{k,i}$ at the server is no smaller than a pre-defined transmit rate $u>0$. The throughput is given by the Shannon's capacity formula, yielding $R_{k,i} = W \log_2(1 + \gamma_{k,i})$ (bits/sec), where $W$ is the bandwidth, and $\gamma_{k,i}$ denotes the signal-to-interference-plus-noise ratio (SINR).

Precisely, the received signal at the server, denoted by $\mathbf{y}_k$, is given by:
\begin{equation}
\mathbf{y}_{k} = \mathbf{h}_{k}^H \mathbf{x}_k + \mathbf{n}_k =  \mathbf{h}_{k}^H  \sum^S_{i=1}\nolimits\mathbf{s}_{k,i} + \mathbf{n}_k,
\end{equation}
where $\mathbf{h}_k$ is the channel coefficient, $\mathbf{n}_k \sim \mathcal{CN}(0,\sigma^2_{\mathbf{h}})$ represents the additive white Gaussian noise (AWGN), and $(\cdot)^H$ stands for the Hermitian transpose operator. According to SD, the SINR for the $i$-th message $s_{k,i}$ is given by:
\begin{equation}
    \gamma_{k,i} = 
    \frac{|\mathbf{h}^H_k \mathbf{s}_{k,i}|^2}{\sigma^2 + \sum^S_{i'=i+1}|\mathbf{h}^H_k \mathbf{s}_{k,i'}|^2}, 
    \label{eq:SINR_chan}
\end{equation}
where $\sigma^2$ denotes the noise power. For ease of notation, the SINR in (\ref{eq:SINR_chan}) is recast as:
\begin{equation}
    \gamma_{k,i} = \chi_k \cdot d^{-\beta} \cdot P_{i} / (\sigma^2 + P^I_{k,i}), \label{eq:sinr}
\end{equation}
in which $\mathbb{E}[|\mathbf{h}_k|^2] = d^{-\beta}$ and $\chi_k \sim \textsf{Exp}(1)$ follow from the large-scale fading and Rayleigh small-scale fading, respectively. The term $P_{k,i}^I = \chi_k d^{-\beta} \hat{P}_i^I$ is the interference, where $\hat{P}_i^I\triangleq\sum^{S}_{i' = i+1}P_{i'}$ for $i\leq S-1$, and $\hat{P}^I_S=P^I_{k,S}=0$ as there is no interference for the last message. 

In SD, the decoding success of the $i$-th message implies that not only the $i$-th but also its all subsequent decoding successes; in other words, $R_{k,1}>u, R_{k,2}>u, ... R_{k,i}>u$ or equivalently $\max\{R_{k,1}, R_{k,2}, ... R_{k,i}\}>u$. Recall that $S$ messages encoded as $\mathbf{x}_k$ experience the same channel. Then, the decoding success probability $p_i$ of the $i$-th message is given as follows:
\begin{align} 
    p_{i} 
    &= \Pr\!\left( \chi_k \!\geq\!  \max \!\left\{\!
    \frac{c}{ P_1/u' - \hat{P}^I_{1}} , \cdots\!, \frac{c}{ P_i/u' - \hat{P}^I_{i}} 
    \!\right\}\!\right)\! \label{eq:sicdsp1} \\
    &= \exp\!\left( -  \max \!\left\{\!
    \frac{c}{ P_1/u' - \hat{P}^I_{1}} , \cdots\!, \frac{c}{ P_i/u' - \hat{P}^I_{i}} 
    \!\right\}\!\right)\!, \label{eq:sicdsp2}
\end{align}
where $c=\sigma^2 d^\beta$, $u' = 2^{\frac{u}{W}}-1$, and the last step follows from the complementary cumulative distribution function (CCDF) of $\chi_k$.

\begin{algorithm}[t!]
\small
Initialize train parameters $\Theta = \{\theta^1,\cdots,\theta^k,\cdots,\theta^K,\theta^G\}$.\\
Split dataset $\bm{Z}$ into $K$ datasets $\bm{Z}=\{Z_1,\cdots, Z_k, \cdots, Z_K\}$.\\
\While{Training}{
    \texttt{//Local Model Training (Algorithm 1)}\\
    \For{$k=1,\cdots,K$}
        {
        \For{$\zeta_k$ in $Z_k$}
            {
            Update local model parameter $\theta^k$ \Comment{Eq.~(\ref{eq:SUSTrain})}
            }
        }

    \texttt{//SC\&SD-based Server Aggregation (Uplink)}\\
    \If{Aggregation Period}{
        $n_\mathsf{L}=|\mathsf{H\cup}\mathsf{F}| \leftarrow 0, n_\mathsf{R}=|\mathsf{F}| \leftarrow 0$, \\ $\mathsf{H} \leftarrow \varnothing, \mathsf{F} \leftarrow  \varnothing$\\
        \For{$k=1,\cdots,K$}
            {$\rho_k \leftarrow rand(1)$\\
            \If{$p_2 \leq \rho_k < p_1$}
                {
                $ \mathsf{H} \leftarrow \mathsf{H} \cup k$,~$n_\mathsf{L} \leftarrow n_\mathsf{L} + 1$
                }            
            \If{$\rho_k \geq p_2$}
                {
                $ \mathsf{F} \leftarrow \mathsf{F} \cup k$,~$n_\mathsf{L} \leftarrow n_\mathsf{L} + 1$,~$n_\mathsf{R} \leftarrow n_\mathsf{R} + 1$
                }
            }
            $\triangleright$ \bf{Case1.} $n_\mathsf{L}>0$, $n_\mathsf{R}>0$\\
            \hspace{10pt}$\theta^G \leftarrow \frac{1}{|\mathsf{H}\cup\mathsf{F}| }{\sum_{k\in\mathsf{H}\cup\mathsf{F}}\theta^k\odot\Xi} + \frac{1}{|\mathsf{F}|}{\sum_{k\in\mathsf{F}}\theta^k\odot\Xi^{-1}} $ \\
            $\triangleright$ \bf{Case2.} $n_\mathsf{L}>0$, $n_\mathsf{R}=0$\\
            \hspace{10pt}$\theta^G \leftarrow \frac{1}{n_\mathsf{L}}\sum_{k\in\mathsf{H}}(\theta^k\odot\Xi)$\\
            $\triangleright$ \bf{Case3.} $n_\mathsf{L}=n_\mathsf{R}=0$\\
            \hspace{10pt} {Skip aggregation }
        }
    \texttt{//Local Update (Downlink)}\\
        \For{$n=1,\cdots,K$}
            {
                $\theta^k \leftarrow \theta^G$
            }
}
\caption{SlimFL with SC \& SD}
\label{alg:SlimFL}
\end{algorithm}

\subsection{SlimFL Operations}
We discuss SlimFL and global model aggregation in further detail. SlimFL is denoted by the symbols shown in Table~\ref{tab:notation-convergence}. \textbf{Algorithm~\ref{alg:SlimFL}} describes the fundamental SlimFL operations. The network consists of $K$ devices that are linked through wireless connections to a parameter server. Each device uses SC whilst communicating with the server, whereas the server uses SD. To be more descriptive, $k$-th device has a local dataset $Z^k \in \bm{Z}$ and an SNN parameter $\theta^k$ with two width configurations. Across devices, the global data $\bm{Z}$ may be IID or non-IID. Each SNN $\theta^k$ is subdivided into an LH $\theta^k\odot\Xi$ and a RH segment $\theta^k\odot\Xi^{-1}$, where $\Xi=\Xi_1$ and $\Xi^{-1}=\Xi_2 - \Xi_1$.
Superposition training is used to train the $k$-th local device (lines 4--9), which is expressed as (\ref{eq:SUSTrain}).
The local device transmits to the server the SC-encoded local model $\theta^k$. 
Each local device transmits two messages (i.e., LH and RH segments), each with a distinct transmission power $P_1$ and $P_2$ relative to $P_1\gg P_2$. 
The uniform random variable $\rho \in [0,1]$ is sampled for user $k$ as $\rho_k$. The sampled $\rho_k$ and the decoding success probability $p_{1}$, $p_{2}$ are used as a criterion for whether LH and RH are successfully decoded. It is identical to calculate (\ref{eq:sicdsp2}).
In accordance with (\ref{eq:sicdsp2}), after reception, the server can successively decode using SD and obtain: (i) a 0.5x model if $\chi \geq  c/(P_1/u' - P_2)$ (lines 15--17) is satisfied; (ii) $1.0$x model if the channel fading gain satisfies $\chi \geq  \max\{ c/(P_1/u' - P_2), c/(P_2/u') \}$ (lines 18--20); and (iii) otherwise it obtains no model. As a result, the RH segments from $\mathsf{F}$ of devices and the LH segments from $\mathsf{H}\cup \mathsf{F}$ of devices are combined by the server.

\section{SlimFL Convergence Analysis}\label{sec:5}
\subsection{Assumptions}
In order to analyze the convergence rate of SlimFL, the following assumptions are considered.
\begin{enumerate}
    \item Regardless of SC or SD, downlink decoding is always successful (\textbf{Algorithm~\ref{alg:SlimFL}}, lines 29--32). The fact that the server (e.g., a base station) has a far higher broadcast power than the uplink power contributes to this.
    \item  We assume that $K$ is big enough that $|\mathsf{H} \cup \mathsf{F}| \approx K p_1 $ and $|\mathsf{F}| \approx K p_2$, where $p_1$ and $p_2$ are the LH and RH segment decoding success probability, respectively, provided in (\ref{eq:sicdsp2}). As a result, during the $t$-th communication cycle, the server builds the following global model $\theta^{G}_{t}$:
    \begin{equation}\theta^{G}_{t} \leftarrow \frac{1}{K p_1}\sum_{k \in \mathsf{H} \cup \mathsf{F}}\nolimits \theta^{k}_{t} \odot \Xi + \frac{1}{K p_2}\sum_{k \in \mathsf{F}}\nolimits\theta^{k}_{t} \odot \Xi^{-1}.
    \label{eq:global}
    \end{equation}
    \item One communication round is assumed per local iteration or mathematically tractability. According to~\cite{ICLR2020FEDMA}, FedAvg does not guarantee that a number of local iterations is proportional to the performance. Since the number of local iterations at each communication round is $1$, superscript $G$ is omitted, resulting in $\theta_t = \theta_t^G$. This assumption will be further discussed in Sec~\ref{sec:6-D-3}.
\end{enumerate}
Based on these assumptions, the convergence analysis of SlimFL is mathematically tractable as discussed in Sec.~\ref{sec:analysis}.

\subsection{Convergence Analysis}\label{sec:analysis}
This paper analyzes the convergence of SlimFL in non-IID data distribution. 
We follow the fundamental derivation techniques utilized \cite{li2019convergence,khaled2020tighter} for FedAvg. %SlimFL convergence analysis, however, is not trivial. One significant reason is because 
One significant challenge in the convergence analysis for SlimFL is due to the local model updates in (\ref{eq:SUSTrain}) and the global model aggregation in (\ref{eq:global}) include sophisticated binary masks owing to the SNN architecture as well as SC and SD. 
As a result, unlike FedAvg, whose global objective function is the weighted average of local loss functions $\{F^k(\theta_t^k)\}$, i.e., empirical risk, SlimFL's objective function $F(\theta_t)$ is unknown. 
Alternatively, we define $F(\theta_t)$ in terms of its gradient $f_t = \nabla F(\theta_t)$, which can be obtained using SlimFL's local and global operations, as discussed below.
After a downlink, the device $k$ replaces its local model with the downloaded global model, represented by $\theta^{k}_{t} \leftarrow \theta_{t}$. The device then changes the local model as:
\begin{equation}
     \theta^{k}_{t+1} \leftarrow \theta_{t} -\eta_{t}g^{k}_{t},
     \label{eq:22}
 \end{equation}
 where $g^{k}_{t} = \sum^{2}_{i = 1} w_{i}\nabla F^{k}(\theta_{t} \odot \Xi_{i}, \zeta^{k}_{t})$ follows from (\ref{eq:SUSTrain}).
 We assume that the student's soft goal may be approximated by the student's hard objective, i.e.,
\begin{equation}
\hat{F}^k(\theta_{t} \odot \Xi_{i}, \zeta^{k}_{t}) \approx F^{k}(\theta_{t} \odot \Xi_{i}, \zeta^{k}_{t}). 
\end{equation}
Next, after the uplink, the server aggregates the updated local models to form the global model $\theta_{t+1}$. By applying (\ref{eq:22}) to (\ref{eq:global}), the resulting global model is as follows:
\begin{align}
\theta_{t+1} 
&\!=\! \frac{1}{Kp_{1}}\!\!\sum_{k \in \mathsf{H} \cup \mathsf{F}}(\theta_{t} \!-\!\eta_{t}g^{k}_{t})\!\odot\!\Xi \!+\!\frac{1}{Kp_{2}}\!\!\sum_{k\in\mathsf{F}}(\theta_{t} \!-\!\eta_{t}g^{k}_{t})\!\odot\!\Xi^{-1}\nonumber\\
&=\theta_{t}-\eta_{t}\Big(\underbrace{\frac{1}{Kp_{1}}\sum_{k \in \mathsf{H}\cup\mathsf{F}}g^{k}_{t}\odot\Xi + \frac{1}{Kp_{2}}\sum_{k \in \mathsf{F}}g^{k}_{t}\odot\Xi^{-1}}_{:=f_t}\Big),
\label{eqn:globalgrad}
\end{align}
resulting in $f_t$ in (\ref{eqn:globalgrad}), which characterizes $F(\theta_t)$. In (\ref{eqn:globalgrad}), the last step can be obtained from $|\mathsf{H}\cup\mathsf{F}|=Kp_1$, $|\mathsf{F}|=Kp_2$, and $\theta_t = \theta_t \odot (\Xi +  \Xi^{-1})$. 

Hereafter we use the bar notation $\bar{\cdot}$ for the value averaged over $\{\zeta_t^k\}$, and $^*$ for indicating the optimum. For the functions $F$ and $\{F^k\}$, we analyze the following commonly used assumptions to prove convergence of FedAvg under non-IID data distributions \cite{li2019convergence,stich2018local}. According to \cite{Stich2106}, following assumptions are justifiable in SNN architectures. 

\begin{assumption}\label{asm:L-smoothness}\textbf{(L-smoothness)}
	$F$ and $\{F^k\}$ are $L$-smooth, i.e., 
	\begin{equation}
	F^k(\theta_{v})  \leq F^k(\theta_{w}) + (\theta_{v} - \theta_{w})^T \nabla F^k(\theta_{w}) + \frac{L}{2} \| \theta_{v} - \theta_{w}\|^2 
	\end{equation}
	for all $v,w>0$.
\end{assumption}
\begin{assumption}\label{asm:mu-strconvex}\textbf{($\bm{\mu}$-strong convexity)}
	$F$ and $\{F^k\}$ are $\mu$-strong convex:
	i.e., 
	\begin{equation}
	F^k(\theta_{v})  \geq F^k(\theta_{w}) + (\theta_{v} - \theta_{w})^T \nabla F^k(\theta_{w}) + \frac{\mu}{2} \| \theta_{v} - \theta_{w}\|^2 
	\end{equation} 
	for all $v, w>0$. 
\end{assumption}
Since linear sum preserves strong convexity and smoothness, $\Tilde{F}^k$ also has the $\mu$-strong convexity and $L$-smoothness. In addition, we also assume our global model has same properties.
\begin{assumption}\label{asm:noniid-grad-bounded}\textbf{(Bounded local gradient variance)}
The variance of the local gradient $\nabla F^k(\theta^k,\zeta^k_t)$ is bounded within $Z_k$, which is given as
	\begin{equation}
	    \mathbb{E}[\|\nabla F^k(\theta^k,\zeta^k_t)-\nabla \bar{F}^k(\theta)\|^2] \leq \sigma_k^2.
	\end{equation}
\end{assumption}

Inspired by \cite{khaled2020tighter}, we define a factor that measures the non-IIDness of $\bm{Z}$ as:
\begin{equation}
    \delta = \frac{1}{K}\sum_{k=1}^K\nolimits \sigma_k^2.
\end{equation}
Indeed, the variance (over $k$) of the local gradient variance (over $Z_k$) is defined as:
\begin{equation}
    \bar{\sigma} \triangleq \frac{1}{K}\sum_{k=1}^K \nolimits \left(\sigma_k - \frac{1}{K}\sum_{k=1}^K \nolimits \sigma_k \right)^2.
\end{equation}  
This characterizes the data distributions over devices, and so does $\delta$ without loss of generality.

To prove the convergence of SlimFL, we derive the following two lemmas as referred to Appendix~\ref{sec:app_lemma1},~\ref{sec:app_lemm2}.
\begin{lemma}\textbf{(Bounded global gradient variance)} \label{lem--1}
Under Assumption~\ref{asm:noniid-grad-bounded}, the variance of the global gradient $f_t$ is bounded within $\bm{Z}$, which is given as
\begin{equation}\EB\|{f}_t-\bar{f}_t\|^2\leq B
\end{equation} where $B = 4\delta (\frac{1}{p_1}+\frac{1}{p_2})\sum^{2}_{i=1}w^2_{i}$.
\end{lemma}

\begin{lemma}\label{lem---2}\textbf{(Per-round global model progress)} Under  Assumptions~\ref{asm:L-smoothness} and \ref{asm:mu-strconvex} with a learning rate $\eta_t \leq \frac{1}{L}$ , the error between the updated global model and its optimum progress as 
% \label{lem--2} By Lemma~\ref{lem--1}, Assumption~\ref{asm:L-smoothness},~\ref{asm:mu-strconvex}, and , then 
\begin{equation}
    \EB\|\theta_{t+1}-\theta^*\|^2\leq (1-\frac{\mu\eta_t}{2})\EB\|\theta_t-\theta^*\|^2+ \eta^2_t B
\end{equation}
\end{lemma}

By Lemma~\ref{lem--1} and Lemma~\ref{lem---2}, the convergence of SlimFL is proved. The full derivation of this proof is in Appendix~\ref{sec:app_theorem}.
\begin{theorem} (\textbf{SlimFL Convergence})\label{convth2}
Under Assumptions \ref{asm:L-smoothness}--\ref{asm:noniid-grad-bounded} with the learning rate $\eta_t = \frac{2}{\mu{t}+2L-\mu}$, one has

\begin{align}
\EB[F(\theta_{t})] - F^{*} 
\leq \frac{L}{\mu}\cdot\frac{\mu L \Delta_1 + 2B }{\mu t+ 2L-\mu},\label{eq:thm1}
\end{align}
where $B = 4\delta (\frac{1}{p_1}+\frac{1}{p_2})\sum^{2}_{i=1}w^2_{i}$ and $\Delta_t \triangleq \EB\|\theta_t-\theta^*\|^2$. Therefore, $\EB[F(\theta_{t})]$ converges to $F^{*}$ as $t\rightarrow\infty$.
\end{theorem}
\subsection{Several Insights of Convergence Analysis}
The result of Theorem 1 exhibits several insights into SlimFL. 
\subsubsection{Failure under extremely poor channels} 
Consider a channel with very poor quality, where the server is unable to decode 1.0x models and hence collects only 0.5x models (i.e., $p_2\approx 0$ and $p_1>0$). In this case, despite the optimality gap aggregating 0.5x models, the term $B$ diverges. In these channel conditions, SC becomes inefficient, and vanilla FL with 0.5x models is preferred to SlimFL.

\subsubsection{Robustness to poor channels}
In (\ref{eq:thm1}), we confirm that an increasing number of 0.5x and 1.0x models aggregated (i.e., increasing $p_1$ and $p_2$) helps equally to reach the global optimal bound. 
As a result, aggregating 0.5x models can further alleviate 1.0x models' frequent decoding failures on poor channels.

\subsubsection{Robustness to non-IID data} 
The optimality gap widens as $\delta$ (i.e., more non-IID). Contrary to vanilla FL, which benefits primarily from aggregating either 0.5x or 1.0x models, SlimFL's increased gap may be mitigated by aggregating not just 1.0x but also 0.5x models. Hence, we conclude that SlimFL is superior for non-IID data distributions and moderately poor channel conditions where $ 0 \ll p_1, p_2 < 1 $. 
Vanilla FL with just 1.0x models or 0.5x models is recommended for highly excellent (i.e., $ p_2 \approx 1 $) or extremely bad (i.e., $ p_1 \approx  0 $) channel conditions, respectively.
SlimFL's favorable conditions and efficacy will be validated by simulation in Sec.~\ref{sec:6}.

\subsubsection{Design principle of SlimFL}  Additionally, Theorem 1 gives design principles for SC and ST as detailed in the following two propositions.

\begin{proposition}[\textbf{Optimal SC power allocations}]\label{remark:1}
Consider the SC power allocation ratio $\lambda \in (0.5,1]$ such that $P_1=\lambda P$ and $P_2 = (1-\lambda)P$. If $\lambda \gg \max\left\{ 0.5, {c u'(1+u')}/{P} \right\}$, 
the optimal SC power allocation ratio that minimizes the RHS of (\ref{eq:thm1}) is given as
\begin{equation}
\lambda^* = \frac{u'+\sqrt{1+u'} -1}{u'}.
\end{equation}
\begin{proof}
Define $D\triangleq \frac{1}{p_1}+\frac{1}{p_2}$. According to the RHS of (\ref{eq:thm1}), $\lambda^*$ minimize $D$. Since $P_1>P_2$, we have \begin{equation}
    D = \exp\left(-\frac{c}{\lambda P/u'-(1-\lambda )P}\right)+\exp\left(-\frac{c}{ (1-\lambda)P/u'}\right).
\end{equation}
If $\lambda \gg c u'(1+u')/{P}$, 
we can approximate both terms in $D$ using the first-order Taylor expansion, yielding 
\begin{equation}
D\approx 2+ \frac{c}{\lambda P/u'-(1-\lambda )P}+\frac{c}{ (1-\lambda)P/u'}. 
\end{equation}
The approximated $D$ is convex, and the first-order necessary condition gives the optimum.
\end{proof}
\end{proposition}
Note that the condition $\lambda \gg \max\left\{ 0.5, {c u'(1+u')}/{P} \right\}$ can be satisfied under sufficiently small model sizes (e.g., $t'\rightarrow 0$), wide bandwidth (e.g., $W\rightarrow \infty$), good channel conditions (e.g., $\sigma^2\rightarrow 0$), and/or a large total transmit power budget (e.g., $P\rightarrow \infty$).

\begin{proposition}[\textbf{Optimal ST ratio}]\label{prop:superposition}
The optimal ST ratio that minimize the RHS of (\ref{eq:thm1}) are given as $w_1^*=w_2^*=1/2$.
\begin{proof}
The RHS of (\ref{eq:thm1}) is minimized at the minimum of $\sum_{i=1}^2 w_i^2$. By the C-S inequality, we have 
\begin{equation}
\sum^2_{i=1}w_i^2 \geq \frac{1}{2}\left(\sum^2_{i=1} w_i\right)^2. 
\end{equation}The desirable result is obtained by combining the condition $\sum^2_{i=1}w_i = 1$ in (\ref{eq:SUSTrain}) with the equality condition of the AM-GM inequality. The influence of $\lambda^*$ and $w_i^*$ will be demonstrated by simulation in the next section.

\end{proof}
\end{proposition}

\section{Experiments}\label{sec:6}
We investigate the effectiveness and feasibility of SlimFL corresponding to the numerical results of convergence analysis, the robustness to various channel conditions and non-IID data distributions, and computation/communication efficiency.
\subsection{Simulation Settings}

\begin{figure}[t!]
\scriptsize\centering
\begin{tabular}{@{}c@{}c@{}}
    \includegraphics[width=0.5\columnwidth]{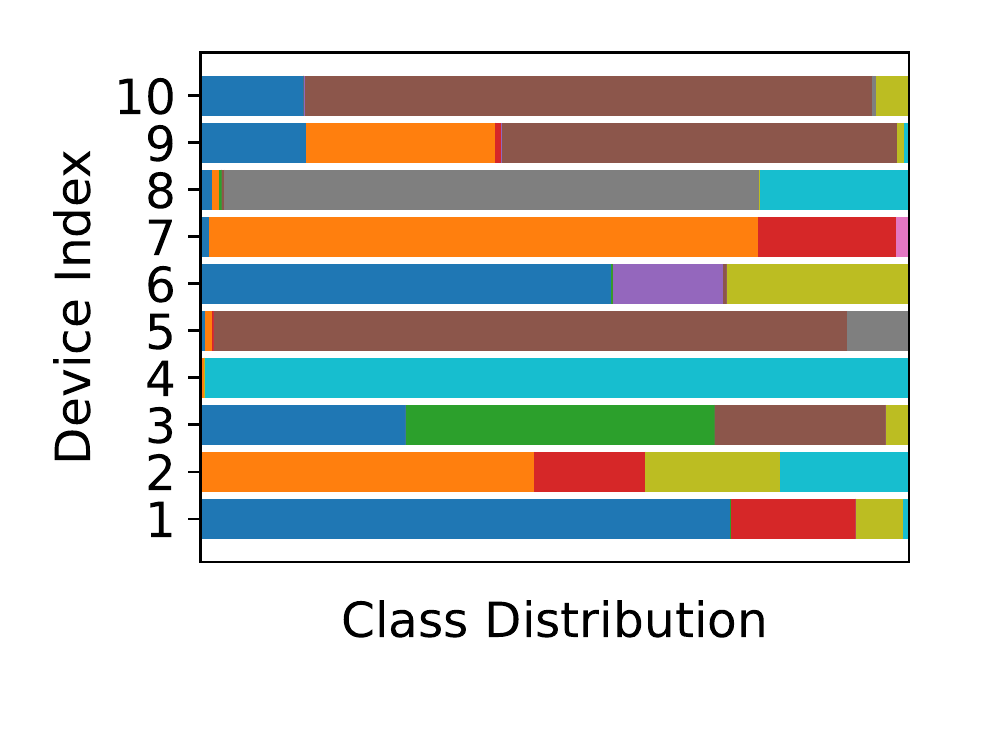} &\includegraphics[width=0.5\columnwidth]{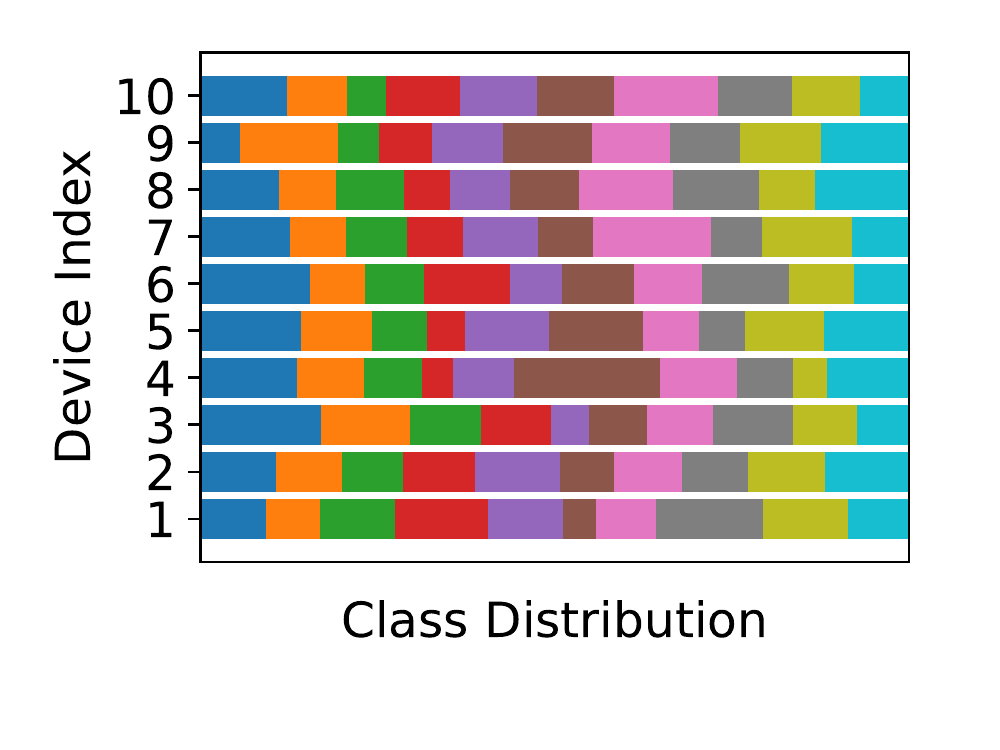} \\
    \small (a) Non-IID ($\alpha = 0.1$). &
    \small (b) IID ($\alpha = 10$). \\ 
\end{tabular}
    \caption{An illustration of the data distributions across $10$ devices for the different values of the Dirichlet concentration ratio $\alpha$. Note that each color represents each class constituting the dataset.}
    \label{fig:Non-iid-distribution}
\end{figure}
To show the effectiveness of the proposed SlimFL, we consider a classification task with the Fashion MNIST (FMNIST) image dataset~\cite{FMNIST}. The impact of other datasets such as the CIFAR-10 and MNIST datasets will be discussed in \ref{sec:Feasibility}. The dataset is randomly sharded and distributed across devices. Following~\cite{Arxiv_2019_NonIID}, the non-IIDness of the data distribution is modeled using a Dirichlet distribution with its concentration parameter $\alpha \in \{0.1, 1.0, 10\}$, where the higher $\alpha$ implies the more non-IID data distribution as illustrated in Fig~\ref{fig:Non-iid-distribution}. For numerical experiments, we adopt the light model of slimmable MobileNet, i.e., UL-MobileNet\footnote{This paper utilizes \textit{Ultra-Light MobileNet} (UL-MobileNet) under the consideration of the limited computing capacity of local.It follows MobileNet architecture~\cite{howard2017mobilenets}. UL-MobileNet consists of five convolution layers activated by the ReLU6 function, average pooling function, and linear layer.}. We consider a single communication round consisting of a pair of uplink and downlink transmissions for each local training iteration. The communication channels over different devices are orthogonal in both uplink and downlink. The Rayleigh fading gain $\chi$ for each channel realization follows an exponential distribution $\chi\sim\textsf{Exp}(1)$~\cite{TseBook:FundamaentalsWC:2005}. The corresponding communication and neural network hyperparameters are summarized in Table~\ref{tab:parameters}. 
We adopt Vanilla FL as a comparison technique. Vanilla FL is widely used FL technique (i.e., FedAvg) with a fixed-width SNN model without leveraging SC nor SD. We consider three Vanilla FL schemes as listed follows:
\begin{enumerate}
    \item \textit{Vanilla-FL $0.5$x/$1.0$x}: Vanilla FL $0.5$x and $1.0$x use only $0.5$x- or $1.0$x-width configurations, respectively. We compare model accuracy and the convergence of SlimFL with Vanilla FL-$0.5$x and $1.0$x corresponding to the robustness to various channel conditions and non-IIDness in Sec.~\ref{sec:performance}.
    \item \textit{Vanilla-FL $1.5$x}: Due to the lack of width-adjustable SNNs, each device in Vanilla FL-$1.5$x separately runs fixed-width $0.5$x and $1.0$x models. Then, devices can choose the $0.5$x or $1.0$x model considering energy heterogeneity, which SlimFL can. In other words, Vanilla FL-$1.5$x operates the two FedAvg operations separately for $0.5$x and $1.0$x models by doubling the bandwidth, transmission power, and computing resources. By comparing SlimFL to Vanilla FL-$1.5$x, we investigate the difference of accuracy, received bits and energy cost.
\end{enumerate}

\begin{table}[t!]
\caption{Simulation Parameters.}
\label{tab:parameters}
%\small
{\small
\begin{center}
	\begin{tabular}{l|r}
    \toprule[1.0pt]
    \bf{Description}  & \bf{Value} \\
    \midrule
        Initial learning rate ($\eta_0$) & $10^{-3}$ \\
        Optimizer                     & Adam         \\ 
        Batch size                     & $64$           \\
        Distance  ($d$)               & $100~\mathrm{[m]}$\\
        Path loss exponent ($\beta$)  & $2.5$ \\
        Bandwidth per device ($W$)& $75~\mathrm{[MHz]}$ \\
        Uplink transmission power  ($P$) & $23~\mathrm{[dBm]}$ \\
        %Downlink transmission power ($P^T_{\mathsf{d}}$) & 100 [mW]\\
        Central frequency  &  $5.9~\mathrm{[GHz]}$ \\
        Noise power spectrum  &  $-169~\mathrm{[dB/Hz]}$\\
    \bottomrule[1.0pt]
	\end{tabular}
\end{center}
}
\end{table}

\subsection{Efficiency, Robustness, and Scalability of SlimFL}\label{sec:performance} 
We carry out to analyze SlimFL's performance to Vanilla FL in situations with a variety of communication conditions and non-IID settings. To assess the efficiency of communication and computation, we first calculate computation cost for UL-MobileNet feed-forwarding~\cite{hernandez2020measuring} and communication cost per one communication round.

\begin{table}[t!]
    \caption{Computing costs and transmission power of UL-MobileNet.}
    \label{tab:tab_cost}
     \centering
    \small\begin{tabular}{c|l|cc}
    \toprule[1pt]
      \multicolumn{2}{c|}{\bf{Description}}  & \bf{1.0x} & \bf{0.5x}  \\ \midrule
     \multirow{3}{*}{Computation}  &  MFLOPS / round & $2.76$ & $0.79$ \\ 
       &  \# of parameters & $4,586$ & $2,293$\\
       &  Bits / round & $172,688$ & $86,344$ \\\midrule
        \multicolumn{2}{c|}{Transmission Power ($P$) [mW]} & $132.1$  & $67.4$ \\ 
       \bottomrule[1pt]    
    \end{tabular}
\end{table}
\begin{table}[t!]
    \caption{Transmission and Computing Costs per Communication Round.}
    \small
    \label{tab:energy}
    \centering
    \begin{tabular}{c|c|c}
        \toprule[1pt]
        {\bf{Metric}} & \bf{SlimFL} & \bf{Vanilla FL-1.5x} \\\midrule
         Communication Cost [mW/Round]           & 199.5           & 399.1   \\
        {Computation Cost [MFLOPS/Epoch]} & {$3.56$} & $3.56$\\\bottomrule[1pt]
    \end{tabular}
\end{table}
\begin{table}[t!]
    \caption{Successfully decoded bits of SlimFL, and Vanilla FL.}
    \label{tab:bits}
    \centering
    \small
    \begin{tabular}{l|c|cc}
        \toprule[1pt]
        \multirow{2}{*}{{\textbf{FL Scheme}}} & \multirow{2}{*}{{\textbf{Decoding}}} &  \multicolumn{2}{c}{\textbf{Channel Condition}}\\
        & & {\bf Good} &  {\bf Poor}\\ \midrule
       \multirow{3}{*}{{\textbf{SlimFL}}} & 0.5x &  1.96  & 18.32\\
        & 1.0x  & 198.45 & 130.10 \\
        & None  & 5.46   & 57.44 \\\midrule
       \multirow{2}{*}{{\textbf{Vanilla FL-0.5x}}} & 0.5x & 102.21 & 93.87 \\
        & None & 0.72 & 9.06\\\midrule
       \multirow{2}{*}{{\textbf{Vanilla FL-1.0x}}} & 1.0x & 200.30 & 144.93\\
        & None & 5.56   & 60.96\\\bottomrule[1pt]
    \end{tabular}
\end{table}

\begin{table}[t!]
    \caption{Total Computation cost and Transmission Power of SlimFL and Vanilla FL-1.5x in Various non-IIDness ($\alpha=0.1,1.0,10)$.}
    \label{tab:energy2}
    \centering
    \footnotesize
    \begin{tabular}{c|c||cc|cc}\toprule[1pt]
        \multirow{2}{*}{\bf{Metric}} &  \multirow{2}{*}{\bf{non-IIDness}} & \multicolumn{2}{c|}{\bf{SlimFL}} & \multicolumn{2}{c}{{\bf Vanilla FL-1.5x}}\\
         &  &  \bf{Good} & \bf{Poor} &  \bf{Good} & \bf{Poor}\\\midrule
        \multirow{3}{*}{\shortstack{Communication \\ Cost [W]}} & \bf{$\alpha=0.1$} & 71.0 & 57.3 & 158.8 & 196.8 \\
         & \bf{$\alpha=1.0$} & 8.5 & 10.4 & 15.8 & 36.7 \\
         & \bf{$\alpha=10$}  & 3.03 & 3.51 & 10.2 & 25.4 \\\midrule
         
        \multirow{3}{*}{\shortstack{Computation \\ Cost [GFLOPS]}} & \bf{$\alpha=0.1$} & 1.27 & 1.02 & 1.88 & 2.41 \\
         & \bf{$\alpha=1.0$} & 0.15 & 0.18 & 0.22 & 0.51\\
         & \bf{$\alpha=10$} & 0.05 & 0.06 & 0.14 & 0.35\\ \bottomrule[1pt]
     \end{tabular}
\end{table}

\begin{figure}[t!]
    \centering
    \begin{tabular}{cc}
        \multicolumn{2}{c}{\includegraphics[width=0.9\columnwidth]{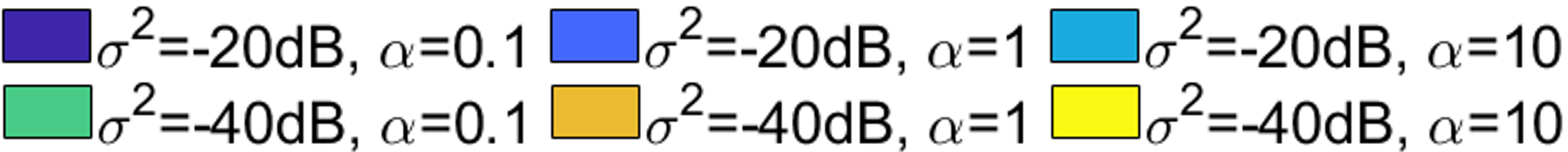}}\\
         \includegraphics[width=0.46\columnwidth]{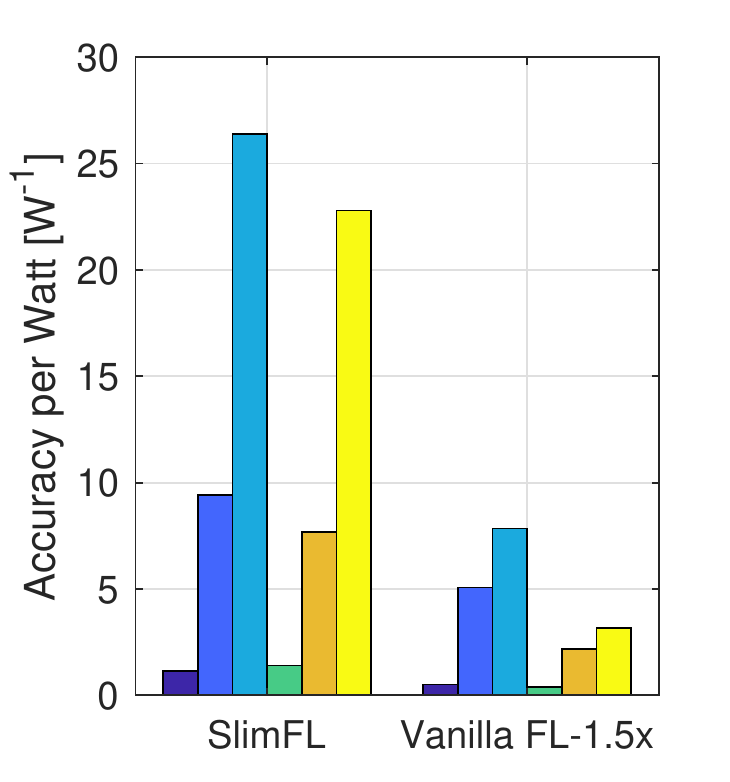} & \includegraphics[width=0.46\columnwidth]{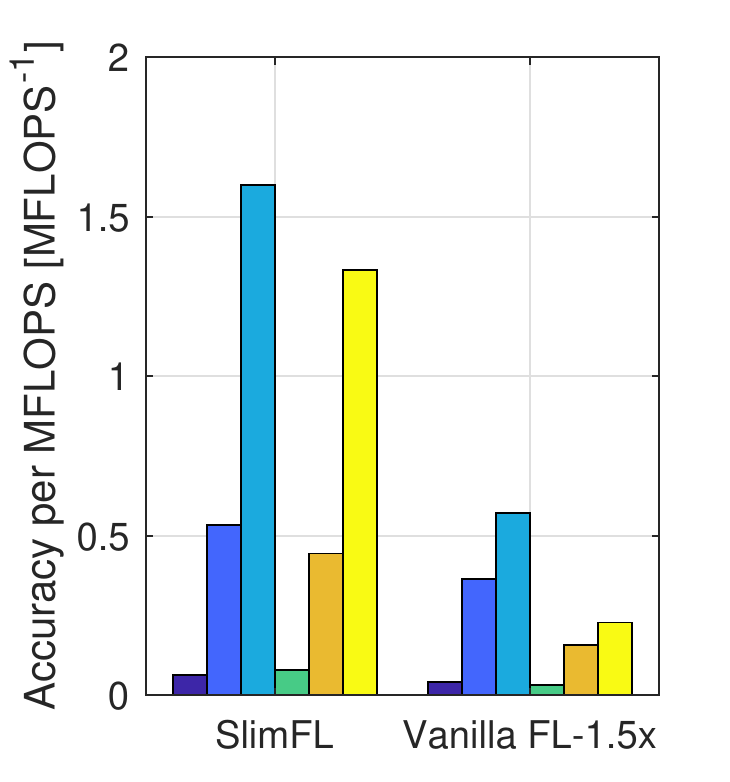}\\
         (a) Communication efficiency. & (b) Computing efficiency.
    \end{tabular}
    \caption{Energy efficiency comparison between SlimFL and Vanilla FL-1.5x, in terms of (a) accuracy per unit communication energy and (b) accuracy per accuracy per unit computing energy.}
    \label{fig:accuracy_per_metric}
\end{figure}
\begin{figure}[t!]
    \centering
    \includegraphics[width=.8\columnwidth]{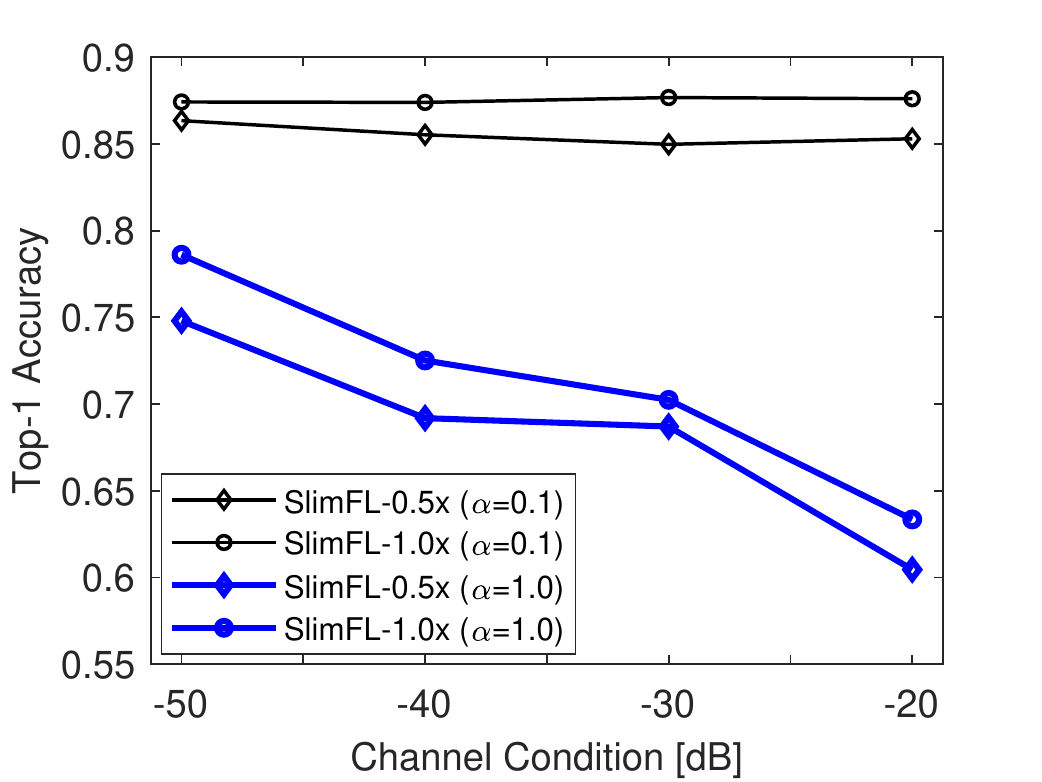}
    \caption{Test accuracy in various channel conditions (e.g., -50dB, -40dB, -30dB and -20dB) with various non-IIDness (e.g., $\alpha=0.1$ and $\alpha=10$).}
    \label{fig:avg-noise-noniid}
\end{figure}

\subsubsection{Communication Efficiency} 
Between ten devices and a server, the total quantity of data communicated is $205.8$MBytes for SlimFL and Vanilla FL-1.0x, and $102.9$MBytes for Vanilla FL-0.5x under ideal channel conditions (i.e., always successful decoding). We investigate the efficiency of communication resources in both good and bad channel conditions to validate the effectiveness of SC and SD. 
The experimental result for this subsection is presented in Table~\ref{tab:bits}, which shows that SlimFL delivers up to $3.52\%$ fewer dropped bits than Vanilla FL-1.0x when SC and SD are being used. 
SlimFL enjoys the benefits mentioned above while consuming only half of the transmission power and bandwidth compared to Vanilla FL-1.5x, as presented in Table~\ref{tab:bits}, corroborating its communication efficiency. Because a part of the transmission power of SlimFL is allocated to 0.5x models, SlimFL decodes less than 1.0x model bits than Vanilla FL-1.0x. In return, SlimFL receives not only 1.0x models but also 0.5x models simultaneously. The additional received 0.5x models coincide with the LH parts of the 1.0x models, which improve the accuracy and convergence speed of both 0.5x and 1.0x models.

\begin{table*}[t!]
    \caption{Accuracy under different channel conditions and~$\alpha$.}
    \label{tab:accuracy}
    \centering
    \small
    \begin{tabular}{c||ccc|ccc}
        \toprule[1pt]
        \multirow{3}{*}{\bf{Method}} &\multicolumn{6}{c}{\bf{Top-1 Accuracy (\%)}} \\ &\multicolumn{3}{c}{\bf{Good}}  & \multicolumn{3}{c}{\bf{Poor}}\\
         & $\alpha=0.1$ & $\alpha=1$ &  $\alpha=10$ & $\alpha=0.1$ & $\alpha=1$ &  $\alpha=10$ \\ \midrule
        SlimFL-0.5x & $54 \pm 2.2$ &  $ 83 \pm 1.0$ & $ 85 \pm 1.0$ & $56 \pm 2.4$ & $82 \pm 1.7$ & $ 85 \pm 1.1$ \\ %--> param 14
        SlimFL-1.0x & $ 59  \pm  2.3$ &  $ 85 \pm 1.1$ & $ 87 \pm 1.1$ & $ 65 \pm 2.9$ & $ 84 \pm 1.4$ & $ 87 \pm 0.9$  \\
        Vanilla FL-0.5x & $ 45 \pm 5.9$ &  $ 84 \pm 1.1$ & $ 85 \pm 1.0$ & $ 39 \pm 8.3$ & $83 \pm 1.2$ & $ 85 \pm 0.9$\\
        Vanilla FL-1.0x & $69  \pm 5.8$ &  $ 85 \pm 4.0$ & $86  \pm  4.3$ & $ 55 \pm 9.2$ & $80 \pm  6.0$ & $ 82 \pm 4.7$ \\ \bottomrule[1pt]
    \end{tabular}
\end{table*}

\begin{figure*}[t!]
\centering
\begin{tabular}{@{}c@{}c@{}c@{}}
    \includegraphics[width=0.321\linewidth]{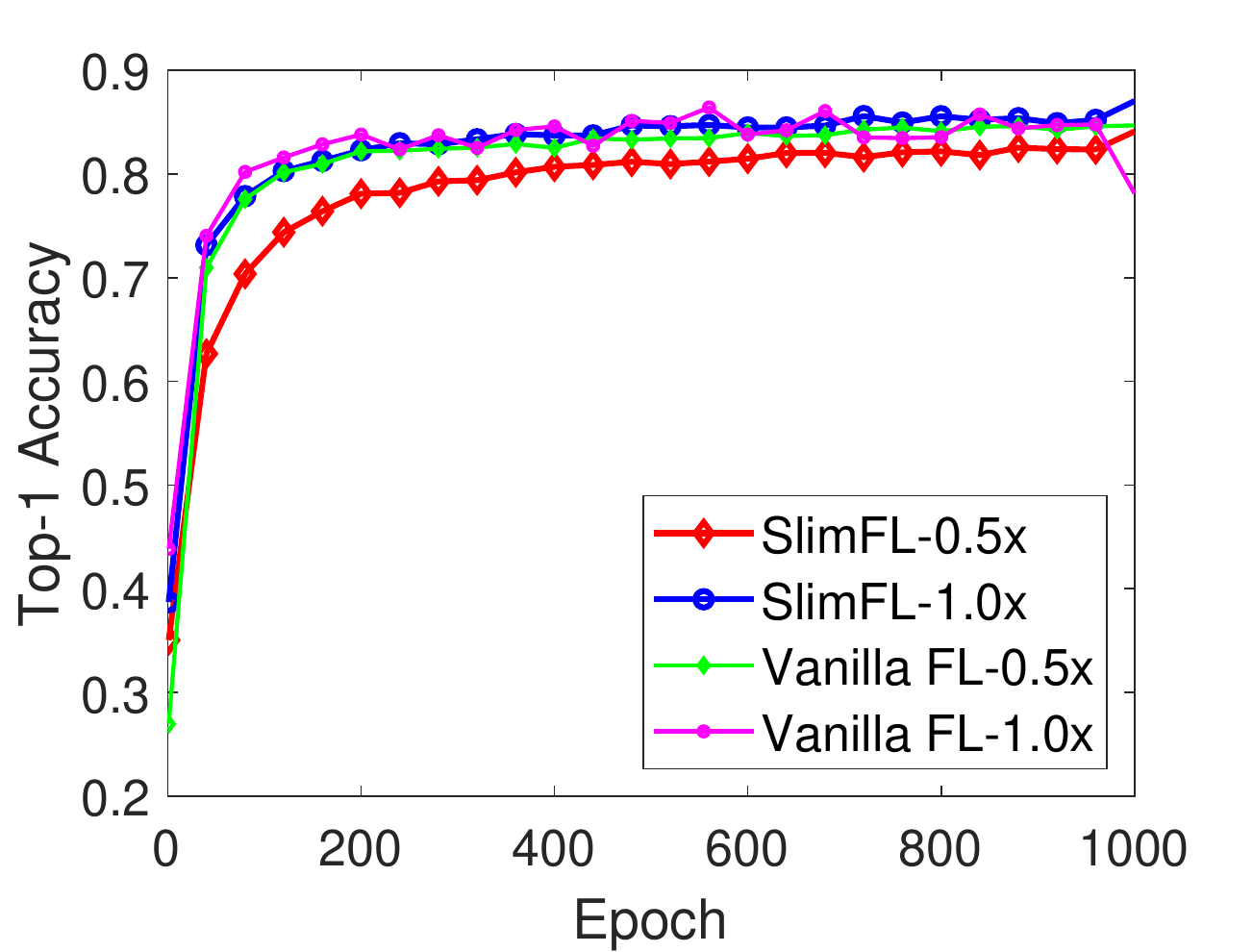}&
    \includegraphics[width=0.321\linewidth]{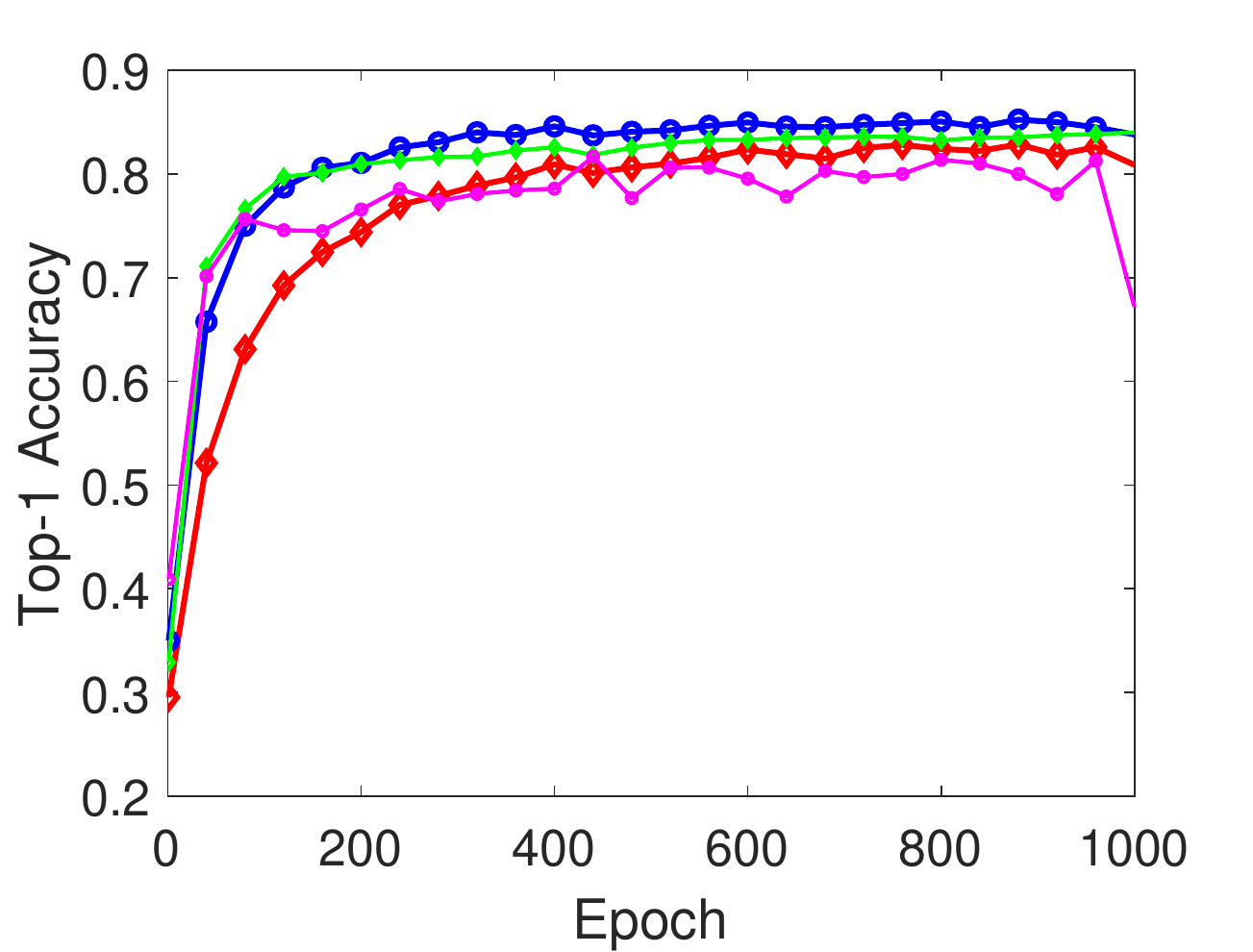}&
    \includegraphics[width=0.321\linewidth]{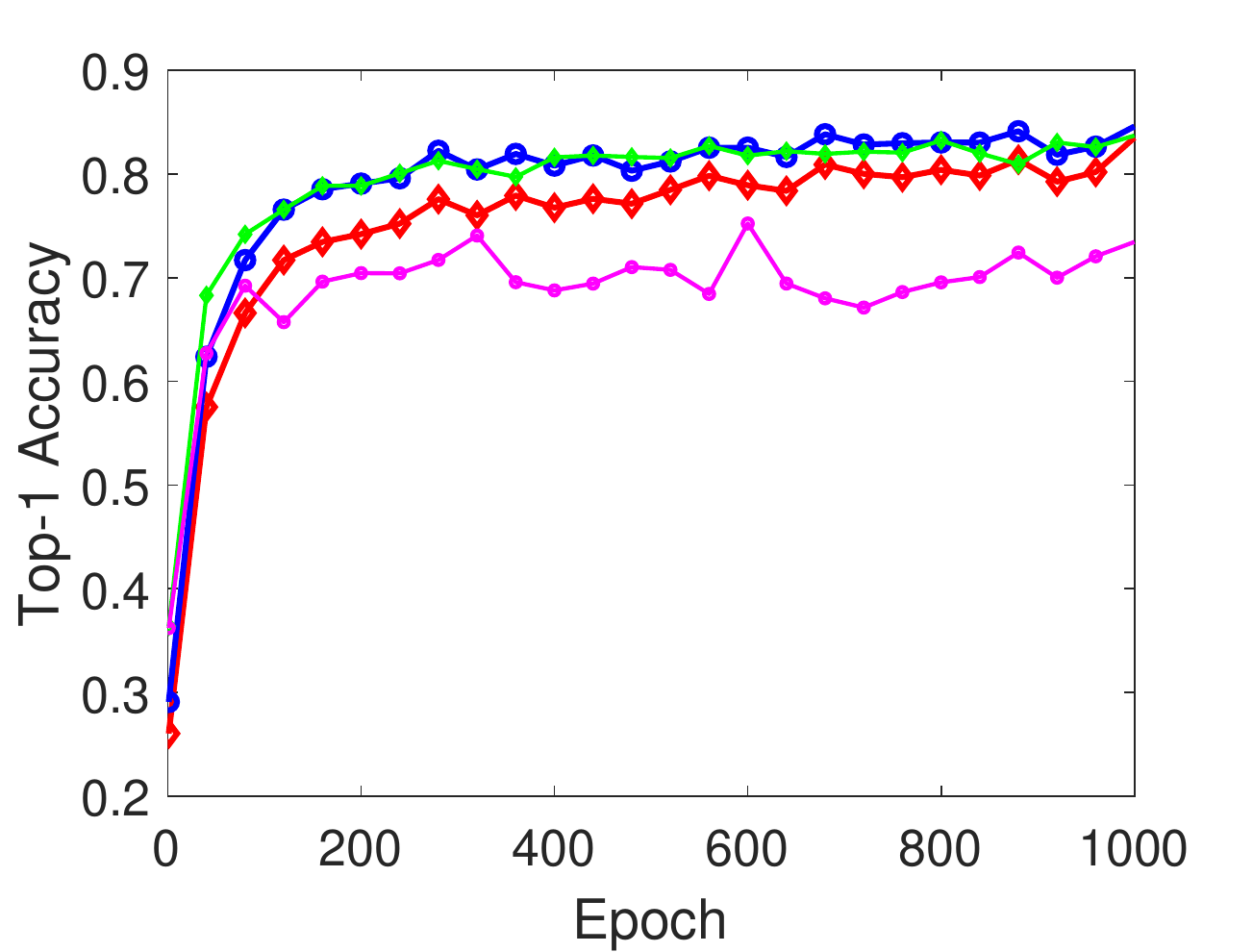}\\
    \small (a) $\sigma^2 = -40\mathrm{dB}$, $\alpha=1$.&
    \small (b) $\sigma^2 = -30\mathrm{dB}$, $\alpha=1$.&
    \small (c) $\sigma^2 = -20\mathrm{dB}$, $\alpha=1$.\\
    \includegraphics[width=0.321\linewidth]{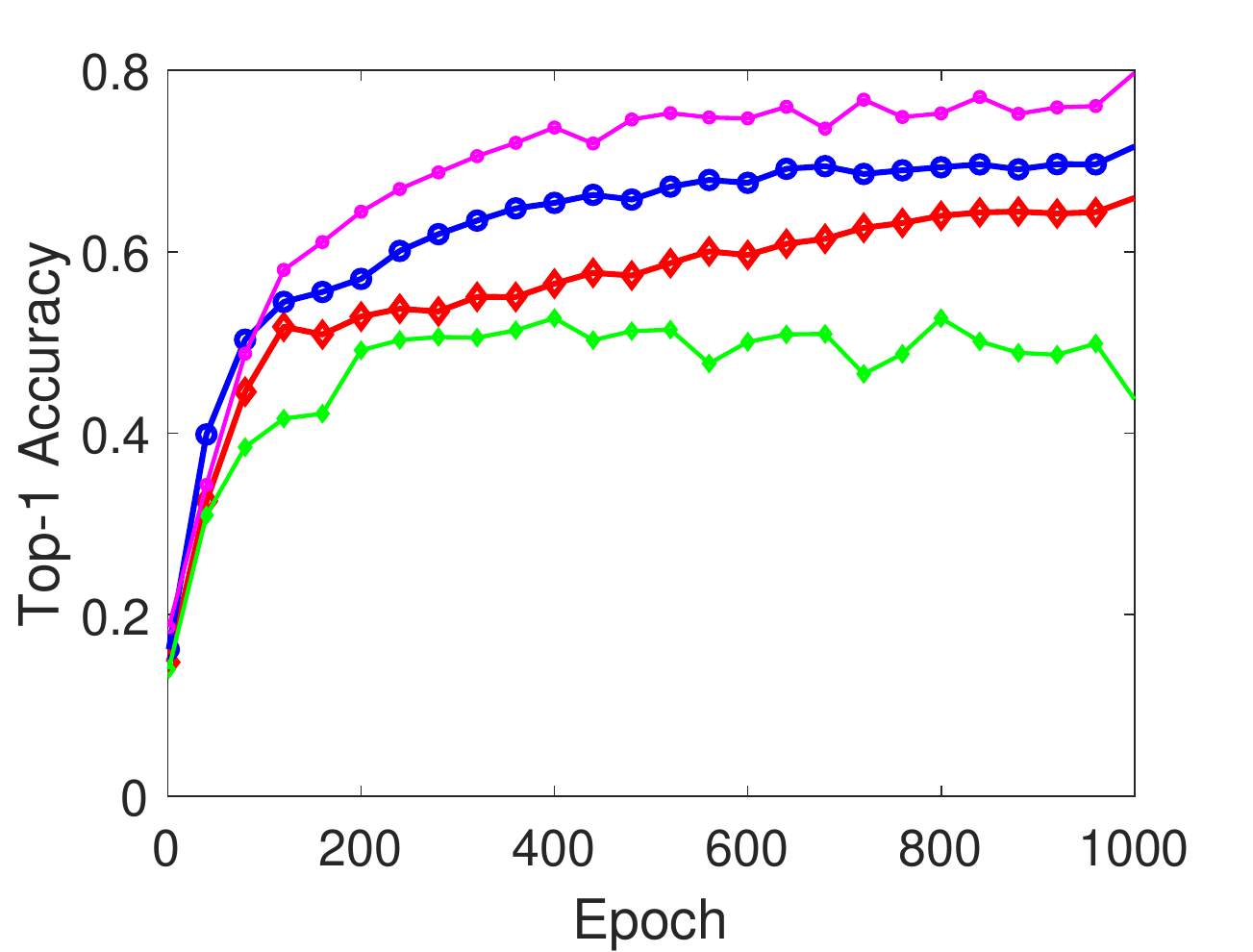}&
    \includegraphics[width=0.321\linewidth]{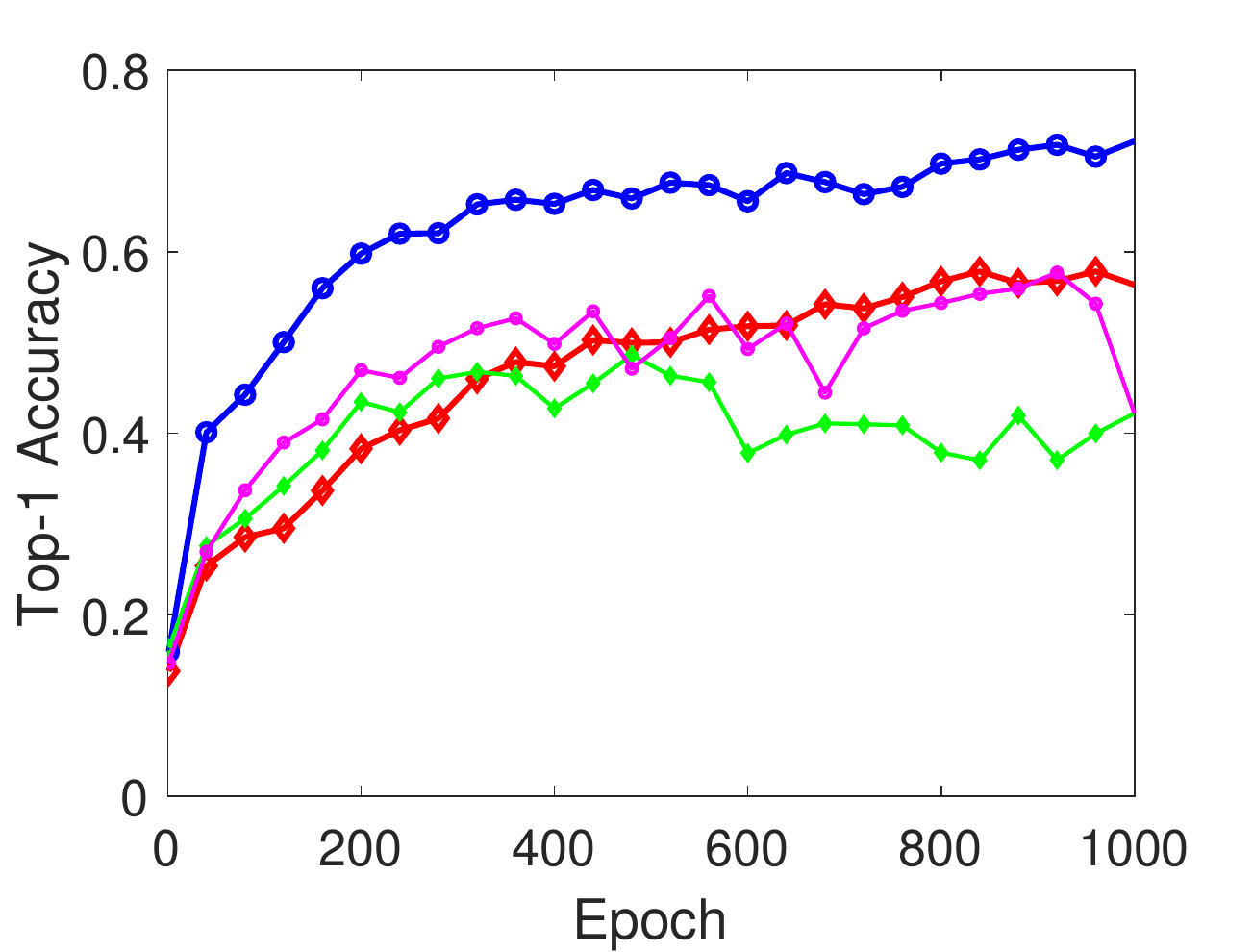}&
    \includegraphics[width=0.321\linewidth]{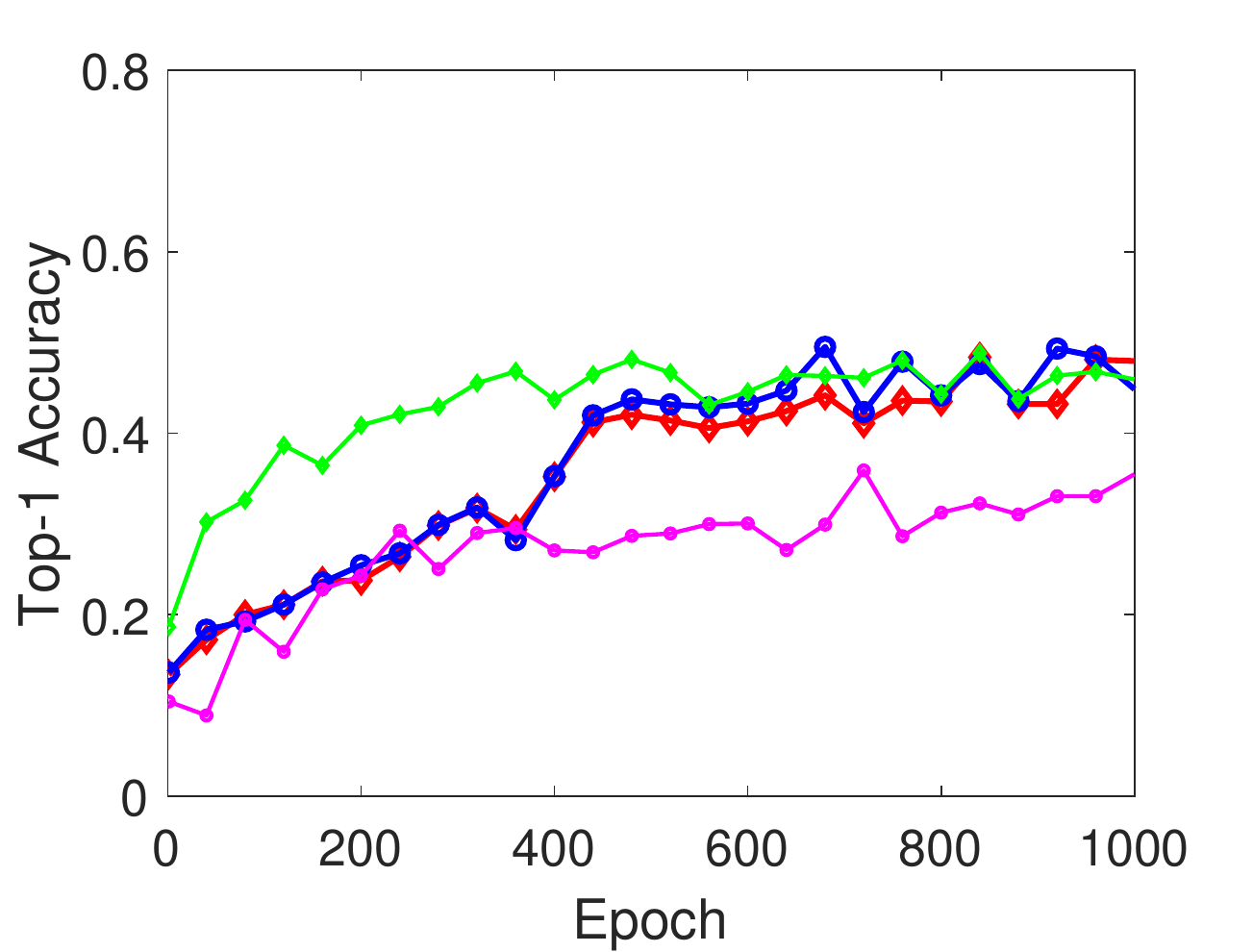}\\
    \small (d) $\sigma^2 = -40\mathrm{dB}$, $\alpha=0.1$.&
    \small (e) $\sigma^2 = -30\mathrm{dB}$, $\alpha=0.1$.&
    \small (f) $\sigma^2 = -20\mathrm{dB}$, $\alpha=0.1$.\\
\end{tabular}
    \caption{{Test accuracy} in various channel noise conditions (on average). (a--c) are with $\alpha=1$, (d--f) are with $\alpha=0.1$.}
    \label{fig:performance-good}
%    \vspace{-3mm}
\end{figure*}
\subsubsection{Energy Efficiency}
Thus far, we have evaluated SlimFL's performance after training using $1,000$ epochs. We estimate the energy consumption till convergence in this section. Numerical convergence in training is defined as the point at which the standard deviation (std) of test accuracy falls below a predefined threshold and the lowest test accuracy exceeds the average test accuracy in $100$ consecutive rounds. To assess model convergence, we establish the reference values for the mean $\mu_{\mathsf{Ref}}$ as $80\%$ and $\sigma_{\mathsf{Ref}}$ as $7.25\%$, respectively. Convergence occurs when the average Top-1 accuracy for $100$ consecutive epochs is greater than $\mu_{\mathsf{Ref}}$ and the average std is less than $\sigma_{\mathsf{Ref}}$. 
%According to \cite{savazzi2021framework},the end-to-end power consumption is defined as follows:
%\begin{align}
%    E = & K \cdot T \cdot (E_C + E_L),\\
%    \mathrm{s.t.~~}  &E_C =L(\theta) \cdot \left(\frac{CI}{EE_U} + \gamma \frac{CI}{EE_D}\right),\\
%     &E_L =  \frac{(\varphi + \beta \gamma ) \cdot CI}{EE_C},
%\end{align}
%, where $E_C, E_L, L(\theta), CI, EE_U, EE_D, EE_C, \varphi, $
% mobile device mflops per watt --> GFLOPs --> watt로 환산 가능 
% 지금 cost reduction은 watt단위로 나와있고, 이를 joule 단위로 (즉, carbon footprint)을 보이는 것은 무리인 것 같고.. 정량적으로 수치를 언급하면서 서술을 해볼까.? 
With this convergence criterion, Table~\ref{tab:energy2} compares the overall energy costs of SlimFL and Vanilla FL-1.5x until convergence, based on the communication and processing energy costs per round in Table~\ref{tab:energy}. As shown in Table~\ref{tab:energy2}, harsh non-IIDness ($\alpha=0.1$) with poor channel conditions makes FL convergence hard. SlimFL, on average, produces a $3.6$x reduction in total computing costs and a $2.9$x reduction in total communication costs until their convergences. In addition, Fig.~\ref{fig:accuracy_per_metric} shows the energy efficiency results corresponding to accuracy. As shown in Fig.~\ref{fig:accuracy_per_metric}(a), SlimFL requires only the $32.63\%$ of Vanilla FL-1.5x's total communication cost for model convergence. Similarly, the computing cost for SlimFL is the $44.37\%$ of Vanilla FL-1.5x's as shown in Fig.~\ref{fig:accuracy_per_metric}(b). This increased energy efficiency is according to SlimFL's faster convergence even in non-IID and/or bad channel circumstances caused by SC and SD.

\subsubsection{Robustness to Poor Channels}
SlimFL and Vanilla FL both achieve high accuracy when channel conditions are good, as shown in Fig.~\ref{fig:performance-good} and Table~\ref{tab:accuracy}. When the channel condition deteriorates from good to poor, however, as shown in Fig.~\ref{fig:performance-good}(c,f) and Table~\ref{tab:accuracy}, Vanilla FL1.0x's maximum accuracy at $\alpha=10$ declines from $86$ to $82$ percent. SlimFL-1.0x, on the other hand, retains its maximum accuracy of $87$ percent at $\alpha = 10$ in both good and bad channel conditions. Additionally, at $\alpha = 0.1$, SlimFL-1.0x outperforms Vanilla FL-1.0x by $18\%$ in terms of top-1 accuracy, despite the fact that Vanilla FL-1.0x consumes more communication and computing resources. Additionally, as the channel condition degrades, the std of Vanilla FL-1.0x's top-1 accuracy increases by up to $59\%$, while SlimFL's std increases by just $31\%$. These findings show SlimFL's resistance to poor channels conditions, as well as its resistance to non-IID data distributions (low $\alpha$) and communication efficiency.

\subsubsection{Robustness to Non-IID Data}
SlimFL-0.5x has a stable convergence under $\alpha=0.1$ condition, as indicated in Fig.~\ref{fig:performance-good}(d--f) and Table~\ref{tab:accuracy} In poor channel conditions and with the non-IID distribution ($\alpha =0.1$), Vanilla FL-0.5x and Vanilla FL-1.0x display the std of $8.3$ and $9.2$. On the other hand, SlimFL-0.5x and SlimFL-1.0x both demonstrate the std of $2.4$ and $2.9$ with top-1 accuracy. This propensity persists even when $\alpha = 1$, $\alpha = 10$ are prevalent. SlimFL-1.0x and SlimFL-0.5x have a smaller coefficient of variation than Vanilla FL-1.0x and Vanilla FL-0.5x. This accentuates SlimFL's robustness to non-IID data on poor channels.

\begin{figure}[t!]
\centering
    \includegraphics[width=0.9\columnwidth]{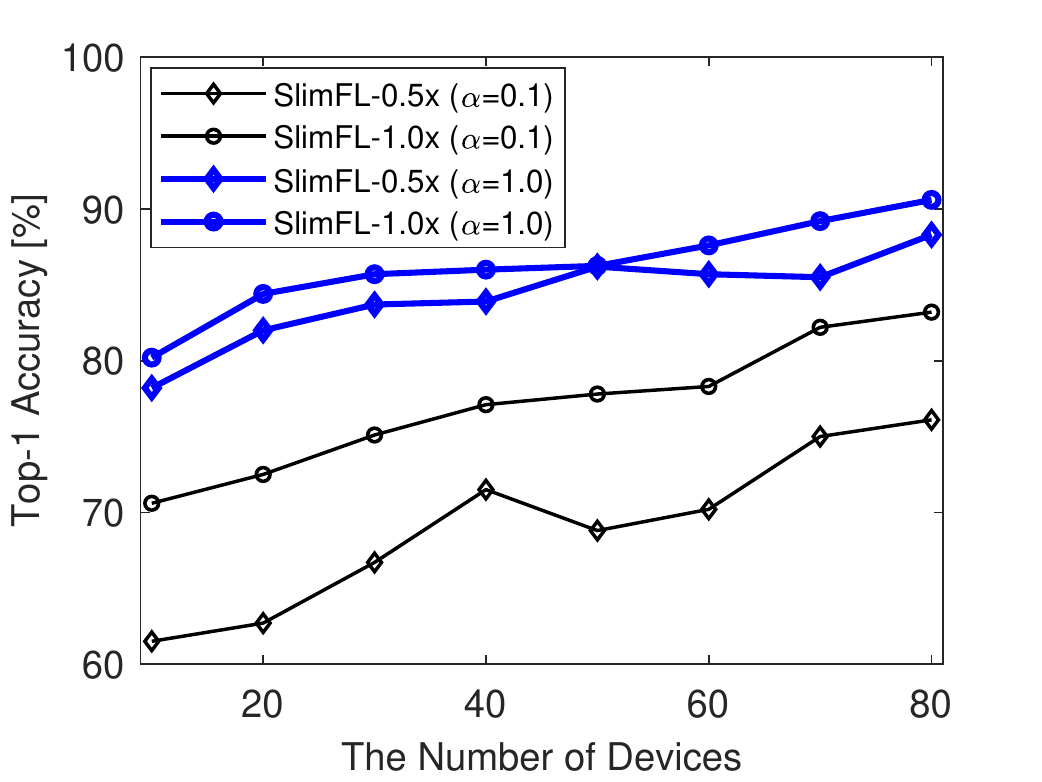}
\caption{Top-1 accuracy with the different number of devices.}
\label{fig:6}
\end{figure}

\subsubsection{Scalability}
As shown in Fig.~\ref{fig:6}, SlimFL's accuracy increases as the number of federating devices increases. The SlimFL-0.5x achieves an accuracy of up to 79\%, while the SlimFL-1.0x reaches an accuracy of up to 85\%. Additionally, with non-IIDness ($\alpha =0.1$) and 70 federating local devices, SlimFL-0.5x achieves a greater level of accuracy than SlimFL-1.0x with 20 federating local devices. Based on the experimental results, it is expected that when constructing a FL system using non-IID datasets, optimality can be obtained by adjusting the number of local devices and width through SlimFL adaption.

\subsection{Ablation Studies on Optimization and Local Training}\label{sec:Ablation}

\begin{figure}[t!]\centering
    \begin{tabular}{c}
        \includegraphics[width=0.9\columnwidth]{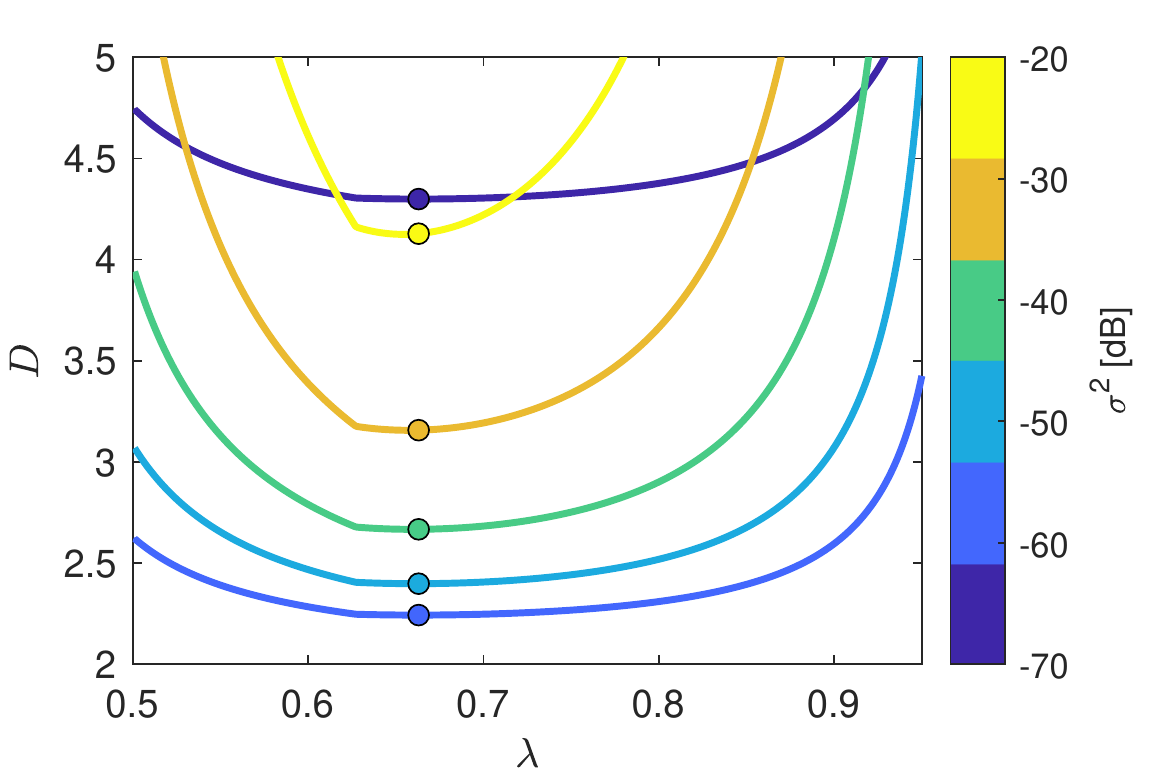}
    \end{tabular}
    \caption{SC power allocation ratio $\lambda$ versus $D$ ($=p_1^{-1}+p_2^{-1}$).}
    \label{fig:allocation}
\end{figure}

\begin{figure}[t!]
\centering
\begin{tabular}{c}
    \includegraphics[width=0.9\columnwidth]{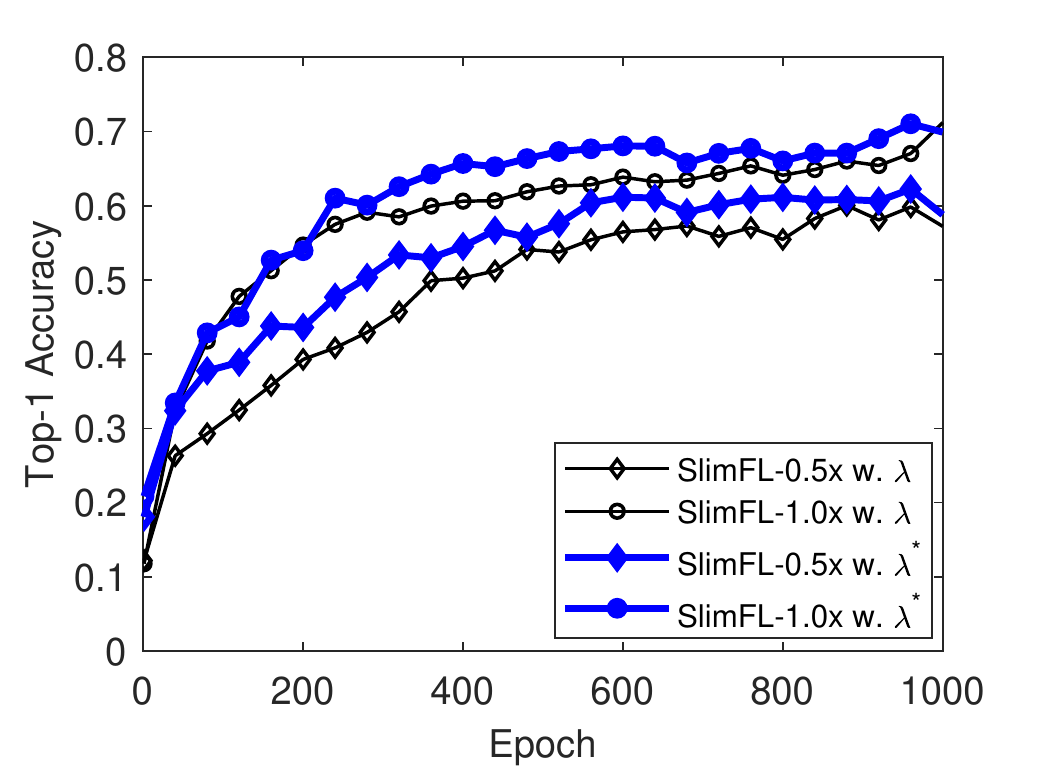}\\
    \small (a) SC power allocation ratio ($\lambda$).\\
    \includegraphics[width=0.9\columnwidth]{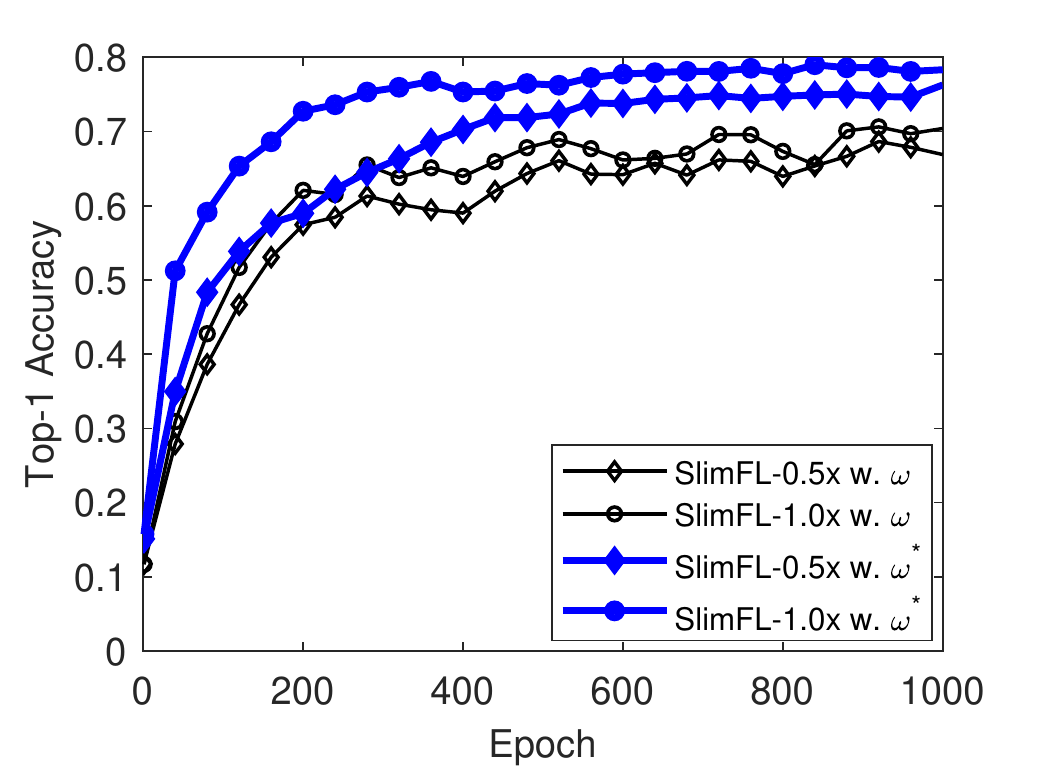}\\
    \small (b) ST ratio ($w_i$)\\
\end{tabular}
\caption{Top-1 accuracy under optimal and non-optimal design parameters: (a) $\lambda^*=0.663$ and $\lambda=0.8$, and (b) $w^*_1=w^*_2 = 0.5$ and ($w_1 = 0.3$, and $w_2 = 0.7$) with $\alpha=0.1$.}
\label{fig:opt}
\end{figure}

\begin{figure}[t!]
\centering
\begin{tabular}{c}
    \includegraphics[width=0.9\columnwidth]{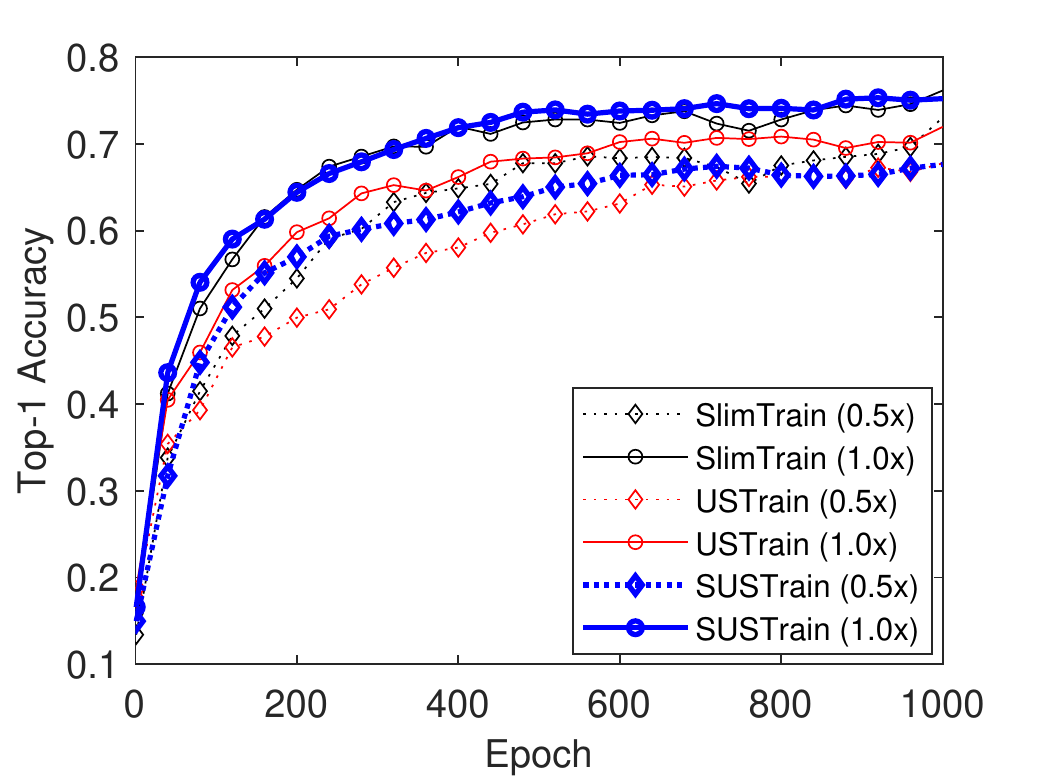} \\
    \small (a) Non-IID ($\alpha=0.1$).\\
    \includegraphics[width=0.9\columnwidth]{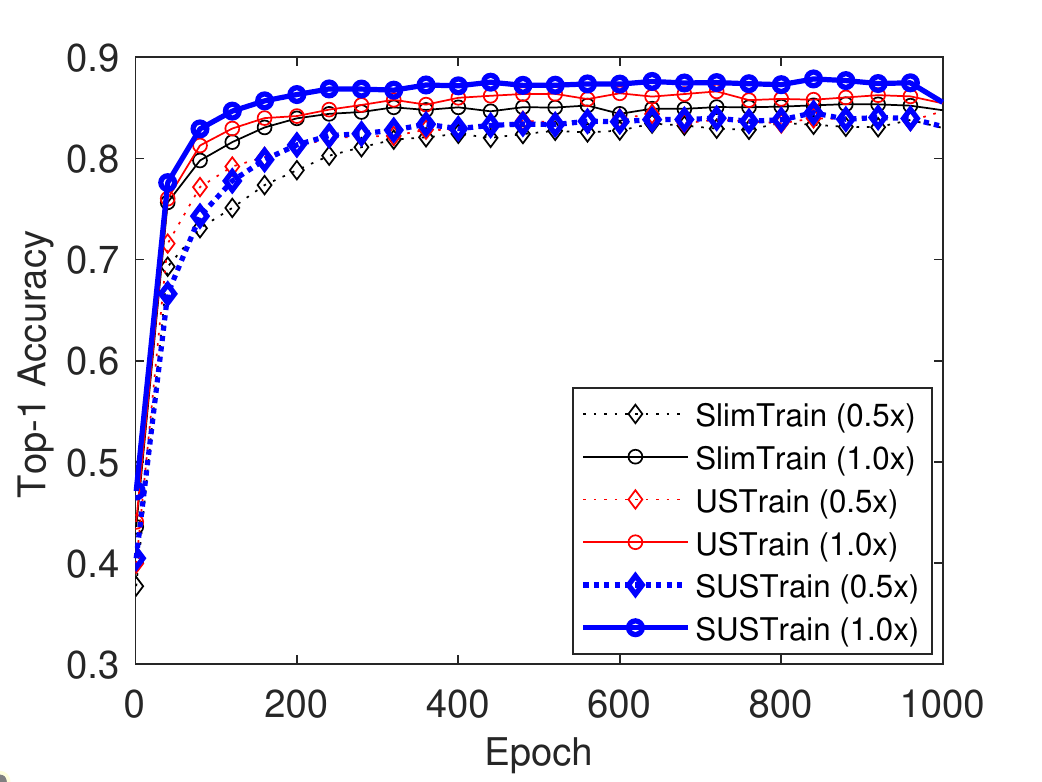}\\
    \small (b) IID ($\alpha=1.0$).
\end{tabular}
\caption{Comparison of SNN training algorithms: SlimTrain \cite{yu2018slimmable}, USTrain \cite{ICCV2019_USlimmable}, and our proposed SUSTrain ($K=10$, $\sigma^2=-30\mathrm{dB}$).}
\label{fig:trainmethod}
\end{figure}

\subsubsection{Effectiveness of SC Power Allocation Optimization}\label{guide:powerallocation}
We present a method for optimizing our suggested model by modifying the ST parameter $w_i$. 
The numerical results of optimal $\lambda^*$ are shown in Fig.~\ref{fig:allocation}. The derivative of the Tayler expansion is used in Proposition~\ref{remark:1} to compute $\lambda^*$. The ideal power allocation factor is $\lambda^*=0.662$, which is the same value as the numerical optimum, as determined by the analytical solution. The simulation was carried out using the baseline with $\lambda=0.8$ in a non-IID settings (i.e., $\alpha = 0.1$) to verify the recommendation from Proposition~\ref{remark:1}. The experiment's result is shown in Fig.~\ref{fig:opt}(a). The top-1 accuracy with $\lambda^*$ is $6.4\%$ higher than top-1 accuracy with 0.5x and $8.8\%$ higher than top-1 accuracy with 1.0x. To put it another way, Proposition~\ref{remark:1} acts as a guiding principle in SlimFL.

\subsubsection{Effectiveness of ST Ratio Optimization}
When $w_1=\cdots=w_S=\frac{1}{S}$ in Proposition~\ref{prop:superposition}, SlimFL has a tight bound. Since $S=2$ is taken into account in our proposed scheme, all hyperparameters that make up ST should be $0.5$, i.e., $(w^*_1,w^*_2)=(0.5,0.5)$. To verify Proposition~\ref{prop:superposition}, we construct a baseline with $(w_1,w_2) = (0.3,0.7)$.  Fig.~\ref{fig:opt}(b) depicts the performance with different $w_i$. Under the optimal ST scheme, top-1 accuracy reaches 78 percent, whereas baseline accuracy is 69 percent. As a result, the ST guideline has a good impact on SlimFL's performance.
 
\subsubsection{Effectiveness of the Proposed Local Training Algorithm} 
To configure out the superiority of the proposed local training algorithm (i.e., SUSTrain), we compare SUSTrain with SlimTrain~\cite{yu2018slimmable} and USTrain~\cite{ICCV2019_USlimmable} which are not only well-known but also state-of-the-art SNN training algorithms. The two training algorithms are presented in \textbf{Algorithm~\ref{alg:slim}} and \textbf{Algorithm~\ref{alg:usslim}}). We investigate the performance between training algorithm in the scheme with ten local devices and $\sigma^2 = -30$dB in two data distributions (i.e., $\alpha = 0.1, 1.0$).
As shown by experiments in Fig. \ref{fig:trainmethod}(a), USTrain is unfit for SlimFL with high non-IID condition (i.e., $\alpha=0.1$), where SlimTrain even outperforms USTrain. With IID data distribution (i.e., $\alpha=1.0$), USTrain outperforms SlimTrain. Fig. \ref{fig:trainmethod} corroborates that regardless of the data distributions, SUSTrain achieves high accuracy with fast convergence, as opposed to USTrain which is effective only under IID data distributions (i.e., $\alpha=1.0$).
We conjecture that the problem comes from the use of outdated teacher's knowledge in USTrain. 
In USTrain, the teacher's logit is set as the value before updating the teacher's model and is compared with a student after updating the teacher's model. Non-IID data distributions exacerbate this mismatch, where the full-width teacher model is significantly updated in the first epoch after downloading the global model due to the huge gap between local and global models.

\begin{table}
\caption{Performance of various width configurations.}
\centering
\small
\begin{tabular}{@{}c|cccccc}
\toprule[1pt]
& \multicolumn{6}{c}{\textbf{SNN}}\\    
\textbf{Metric} & \textbf{1/6x} & \textbf{2/6x} & \textbf{3/6x} & \textbf{4/6x} & \textbf{5/6x} & \textbf{6/6x}\\\midrule
\textbf{Top-1 Accuracy (\%)} & 72.6 & 77.3 & 82.5 & 84.6 & 84.6 &  85.1     \\\midrule
\textbf{Computation Cost per} & \multirow{2}{*}{0.23} & \multirow{2}{*}{0.45} & \multirow{2}{*}{0.97} &  \multirow{2}{*}{1.73} & \multirow{2}{*}{2.71} & \multirow{2}{*}{3.82}     \\
\textbf{Image (MFLOPS)}\\
\bottomrule
\end{tabular}\label{tab:8}
\end{table}

\begin{figure}[t!]
\centering
\includegraphics[width=0.9\columnwidth]{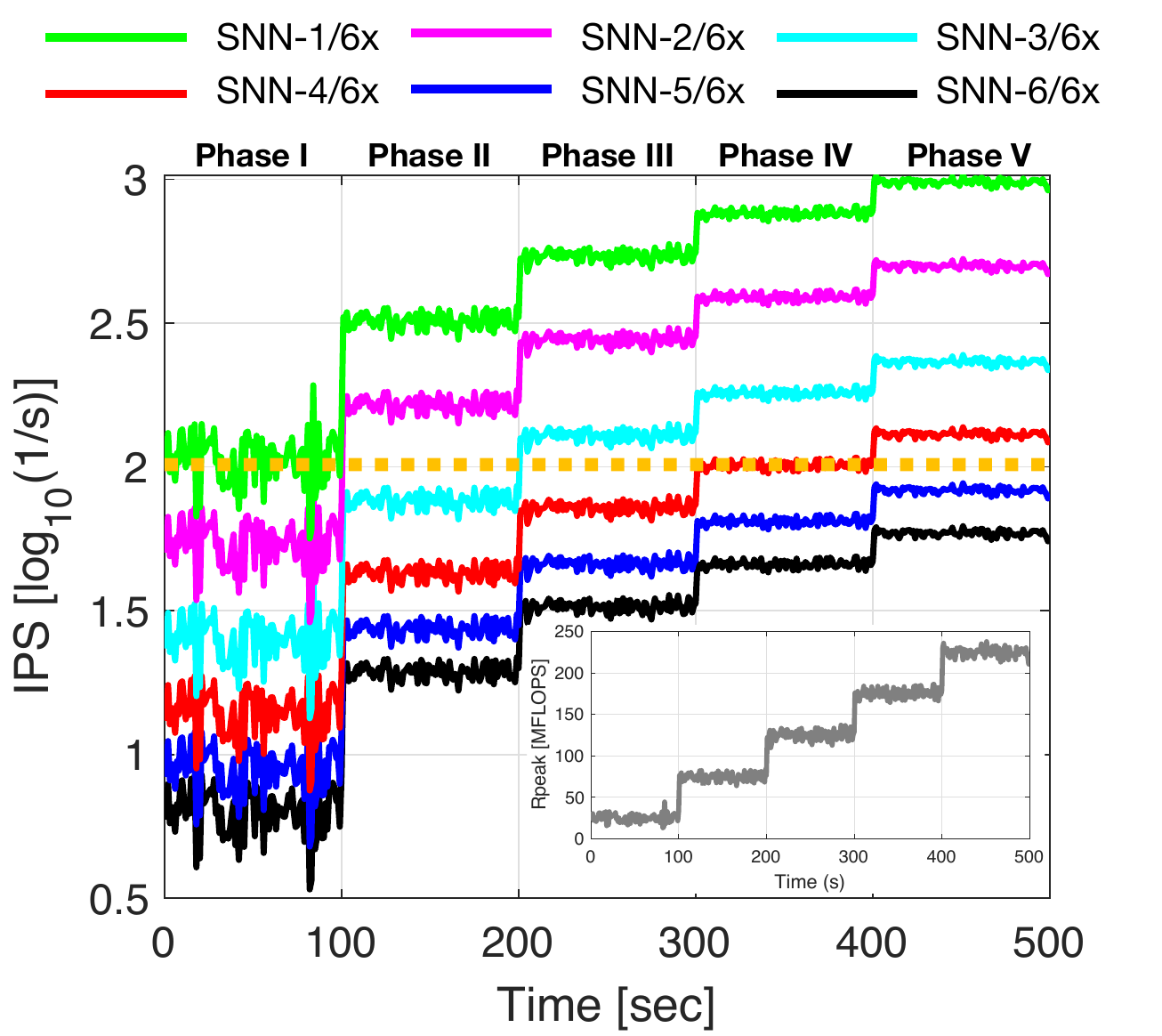}
\caption{Image processing per second of various width configurations given time-varing theoretical CPU resources (i.e., $\textsf{R}_{\textsf{peak}}$).}
\label{fig:hetero}
\end{figure}

\subsection{Impact on Heterogeneous Energy Capacity}
To verify the impact on heterogeneous computing capacities of devices, we investigate six-width configurations (e.g., 1/6x, 2/6x, $\cdots$, and 6/6x). 
First, we train SNNs with the default settings of SlimFL ($\alpha = 10$, $K=10$, $\sigma^2 = -30\text{dB}$). Then, we measure top-1 accuracy and calculate FLOPS. 
Its specification is presented in Table \ref{tab:8}. 
Second, we investigate the dynamic width adaptation to the device, which has a time-varying theoretical peak performance of the CPU, denoted as $\textsf{R}_{\textsf{peak}}$ (MFLOPS). We set $\textsf{R}_{\textsf{peak}} \in [20,230]$ (MFLOPS).  
Lastly, we calculate the number of image processing per second (IPS) by the following metric:
\begin{equation}
\text{IPS} = \frac{\textsf{R}_{\textsf{peak}} \textsf{~[MFLOPS/sec]}}{\textsf{Computational Cost per Image [MFLOPS]}}.
\end{equation}
Fig.~\ref{fig:hetero} shows $\textsf{IPS}$ of multi-width configurations in log-scale, and $\textsf{R}_{\textsf{peak}}$ over time. As shown in Fig.~\ref{fig:hetero}, five different $\textsf{R}_{\textsf{peak}}$ phases exist. Suppose that the mobile devices require 100 images per second, i.e., $\text{IPS}_{\textit{target}}=100$ marked as the \textit{yellow dashed line}. In phase I in Fig.~\ref{fig:hetero}, the IPS requirement is only satisfied when using SNN-1/6x. In phase II, the IPS requirement is satisfied when using SNN-1/6x and SNN-2/6x. However, it is recommended to use SNN-2/6x, because the top-1 accuracy increases 4.7\%. Similarly, SNN-3/6x, SNN-4/6x are recommended to be used in phase III, and phase IV, and V, respectively.
Furthermore, we can use six model configurations with the memory usage of SNN-1.0x. If vanilla NN is used, 3.5x more memory is required than using SNN. Thus, using SNN is an adjustable solution for computational energy heterogeneous conditions.

\subsection{Feasibility Studies on Channel and Training Environments}\label{sec:Feasibility}
We investigate the generalization of SlimFL across various datasets and the different numbers of local iterations in various channel models.

\begin{figure}[t!]
\centering
    \includegraphics[width=0.9\columnwidth]{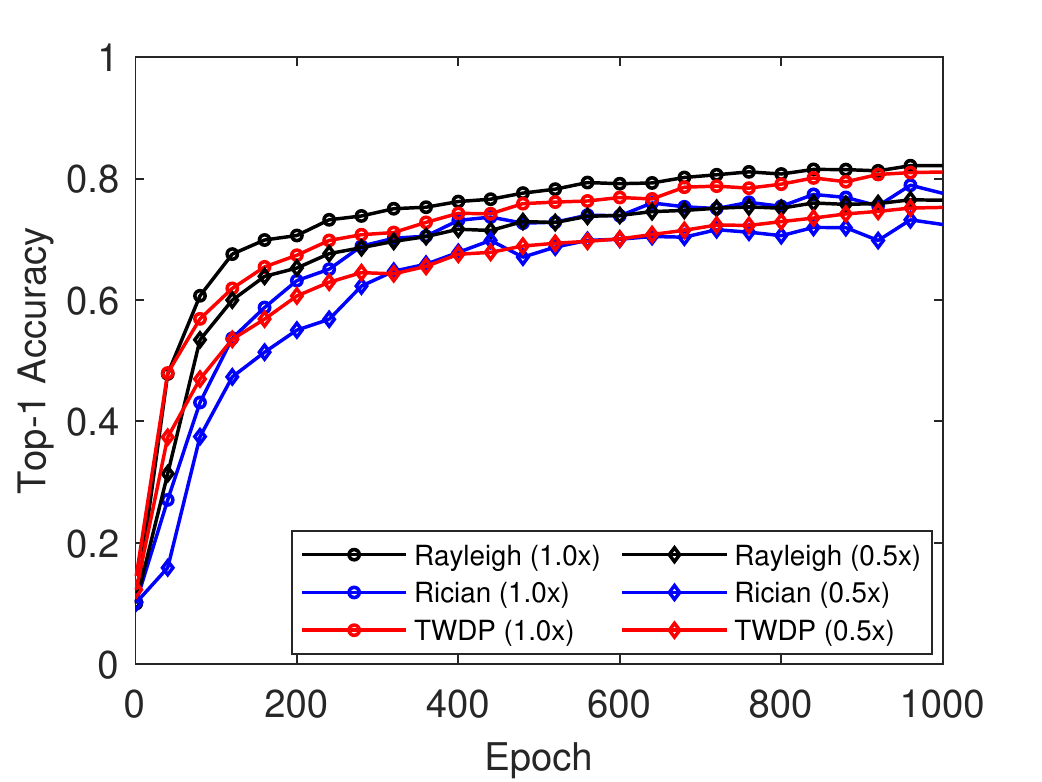}
\caption{Top-1 accuracy with the different fading effects.}
\label{fig:10}
\end{figure} 

\begin{figure}[t!]
\centering
    \includegraphics[width=0.9\columnwidth]{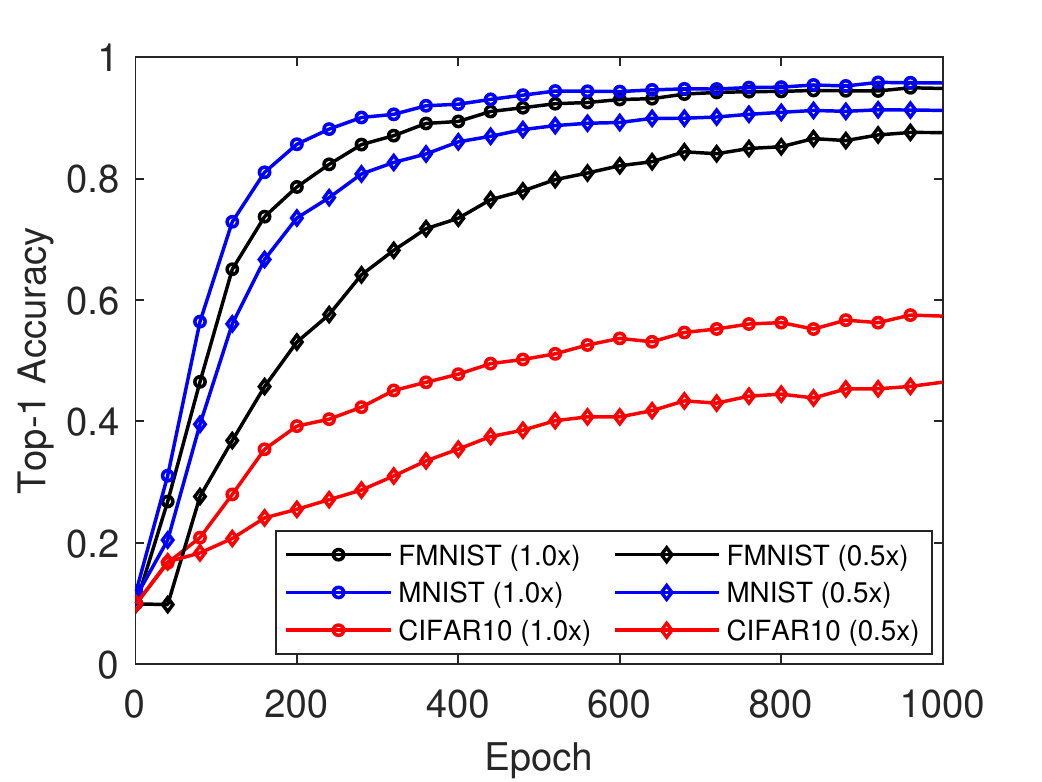}
\caption{Top-1 accuracy with the different datasets.}
\label{fig:11}
\end{figure}

\begin{figure}[t!]
\centering
\begin{tabular}{c}
    \includegraphics[width=0.9\columnwidth]{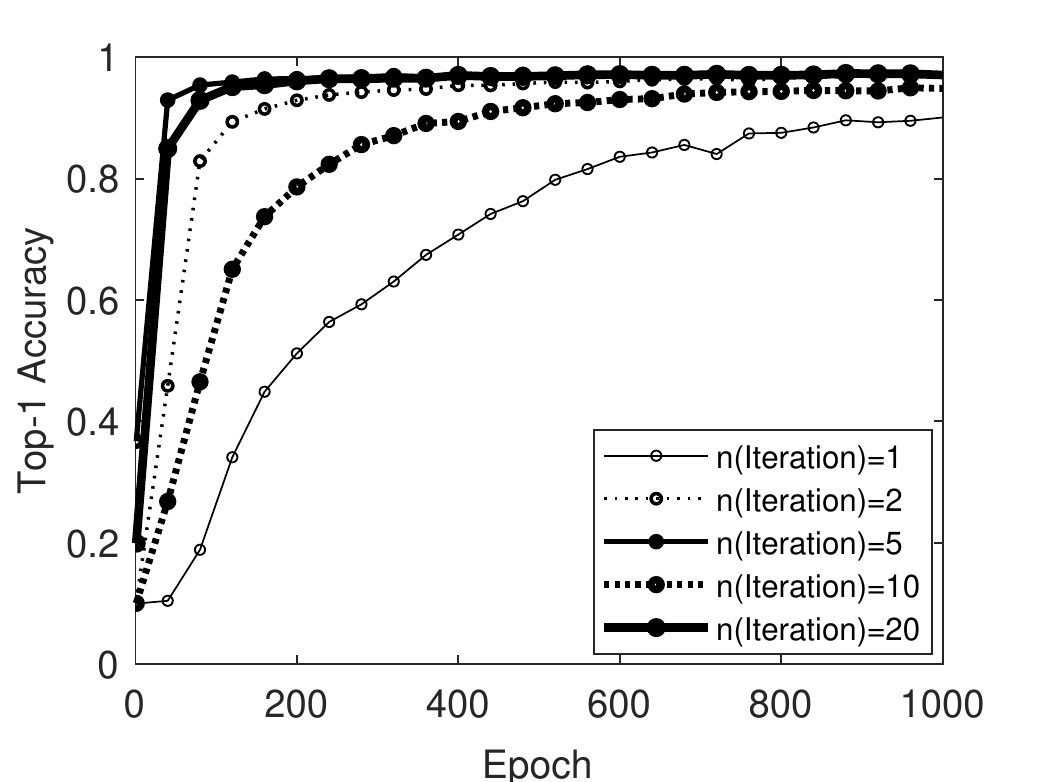} \\
    \small (a) SlimFL-1.0x.\\
    \includegraphics[width=0.9\columnwidth]{fig14a.pdf}\\
    \small (b) SlimFL-0.5x.
\end{tabular}
\caption{Top-1 accuracy with different number of local iterations.}
\label{fig:12}
\end{figure}

\subsubsection{Impact of Channel Models}
We design additional simulations to verify SlimFL works in more realistic channel models (e.g., Rician fading~\cite{Molisch} and Two-wave with diffuse power fading (TWDP)~\cite{1010620}).
In indoor wireless communication, Rician fading is superior than Rayleigh fading. In addition, TWDP is used in air, space or maritime wireless communication, where the channel fading is worse than Rayleigh~\cite{9237116}. We simulate Rician fading with $K$-factor $3.5$, where the fading follows Rician distributions $\chi\sim\mathrm{Rice}(0.5577, 0.2106)$. $(K,\Delta)$-factor of TWDP fading is set to $(3.5, 0.1)$. All communication settings without fading effect are identical to Table~\ref{tab:parameters}. Fig.~\ref{fig:10} is the result of the simulation. As shown in Fig.~\ref{fig:10}, the top-1 accuracy is found to be high in the order of Rayleigh fading, TWDP fading, and Rician fading channel models. SlimFL-1.0x always outperforms SlimFL-0.5x. Thus, SlimFL performs well in a variety of channel models.

\subsubsection{Impact of Datasets}
We conduct additional simulations to verify whether SlimFL operates with other datasets (e.g., CIFAR10~\cite{CIFAR10} and MNIST~\cite{deng2012mnist}). CIFAR10 dataset consists of 60,000 color images with size $32 \times 32$. MNIST is a handwritten digit classification dataset. It consists of 60,000 grayscale images with size $28 \times 28$. Both image datasets have $10$ classes, i.e., digits. CIFAR10 is known to be a more challenging problem, which is harder than MNIST or FMNIST. The simulation is conducted with $\alpha = 1.0, K = 10$, and $\sigma^2 = -30\mathrm{dB}$. Fig.~\ref{fig:11} is the result of the simulation. SlimFL with MNIST dataset shows the best top-1 accuracy, FMNIST data set shows the next highest accuracy, and CIFAR10 shows the lowest accuracy. In terms of the performance between 1.0x and 0.5x, SlimFL-1.0x is always higher than SlimFL-0.5x. In other words, we confirm that our proposed SlimFL works well 
regardless of the dataset.

\subsubsection{Impact of Local Training Iterations}\label{sec:6-D-3}
This paper adopts local iterations per communication round as $1$ for mathematical tractability. Thus, the convergence bound does not contain the local iteration. According to \cite{li2019convergence}, the over-small or over-large local iterations cause high communication and computational cost, while the optimal local iteration exists.
We conduct additional simulations to verify the SlimFL operation under different numbers of local iterations (i.e., $1$, $2$, $5$, $10$, and $20$). As mentioned above, FedAvg does not guarantee that many local iterations are proportional to the performance~\cite{ICLR2020FEDMA}. Similarly in our simulation as shown in Fig.~\ref{fig:12}, the top-1 accuracy of both SlimFL-1.0x and SlimFL-0.5x is high and is on the order of $\{5,20,2,10,1\}$. 
The empirical result shows that there exist at least two local extremums. One is located in $[2,10]$, and the other is in $[5, \infty]$, respectively. In other words, there are at least two local optima corresponding to local iterations in $[1,\infty]$. Note that these local optima do not indicate global optima.

\section{Concluding Remarks}\label{sec:7}
Existing FL solutions are incapable of adjusting flexibly to devices with varying amounts of available energy and channel throughput without jeopardizing communication or energy efficiency.
To overcome this problem, we propose SlimFL, a novel framework for FL over SNNs that utilizes superpositioned training for local SNN training and superposition coding for trained model aggregation. Extensive analyses and simulations show that SlimFL is an energy- and communication-efficient solution in a variety of communication conditions and data distributions. SlimFL achieves higher accuracy and faster convergence while using less energy than vanilla FL, which requires 2X more communication resources. Additionally, studying the impact of more adjustable SNN width levels could be an interesting topic for future work. Another interesting direction is to apply SlimFL for multitask learning with various width configurations in different tasks.

\bibliographystyle{IEEEtran}
\bibliography{ref}
\appendices\section{Proof of Theorem 1}\label{sec:app_lemma}
\subsection{Proof of Lemma 1}\label{sec:app_lemma1}
\begin{proof}
According to $f_t$ in (\ref{eqn:globalgrad}) and Assumption~\ref{asm:noniid-grad-bounded}, 
\begin{multline}
\|f_t-\bar{f}_t\|^2 = \left\|\frac{1}{Kp_1}\sum_{{k}\in\mathsf{H}\cup\mathsf{F}}(g^k_t-\bar{g}^k_t)\odot\Xi \right. \\ 
\left.+\frac{1}{Kp_2}\sum_{{k}\in\mathsf{F}}(g^k_t - \bar{g}^k_t) \odot \Xi^{-1}\right\|^2
\end{multline}
\begin{multline}
\leq \frac{2}{Kp_1}\sum_{k\in\mathsf{H}}\|(g^k_t-\bar{g}^k_t)\odot\Xi\|^2 \\
+ \frac{2}{Kp_2}\sum_{k\in\mathsf{F}}\|(g^k_t-\bar{g}^k_t)
\odot\Xi^{-1}\|^2 
\end{multline}
\begin{equation}\label{eq:f1}
\leq  \frac{2}{Kp_1}\sum_{k \in \mathsf{H}}\|g^k_t-\bar{g}^k_t\|^2 + \frac{2}{Kp_2}\sum_{k\in\mathsf{F}}\|g^k_t-\bar{g}^k_t\|^2, 
\end{equation}
where the first inequality follows from the Cauchy–Schwarz (C-S) inequality, and the last step is because $\|X\odot\Xi\|^2 \leq \|X\|^2$. 
Similarly, we have
\begin{align}
\left\|g^k_t-\bar{g}^k_t\right\|^2 
& = \left\|\sum^2_{i=1} w_i(\nabla F^k(\theta_t,\zeta^k_t)-\nabla F^k(\theta_t))\odot \Xi_i\right\|^2  \nonumber\\
& \leq 2 \sum^2_{i=1} w^2_i\left\|\nabla F^k(\theta_t,\zeta^k_t) - \nabla F^k(\theta_t) \right\|^2.
\end{align}
Taking an expectation at both sides yields 
\begin{equation}\label{eq:f2}
\EB\|g^k_t-\bar{g}^k_t\|^2\leq 2\sigma^2_k\sum^2_{i=1} w^2_i.
\end{equation}
Combining (\ref{eq:f1}) and (\ref{eq:f2}) finalizes the proof.
\end{proof}
\subsection{Proof of Lemma 2}\label{sec:app_lemm2}
\begin{proof}
According to (\ref{eqn:globalgrad}), we have
\begin{equation}
    \|\theta_{t+1}-\theta^*\|^2 = \|\theta_t-\eta_t f_t - \theta^* - \eta_t \bar{f}_t + \eta_t\bar{f}_t\|^2 
\end{equation}
\begin{eqnarray}
    &=&\underbrace{\|\theta_t-\theta^*-\eta_t\bar{f}_t\|^2}_{A_1}\nonumber \\ & & +\underbrace{2\eta_t\langle\theta_t-\theta^*-\eta_t f_t,\bar{f}_t-f_t\rangle}_{A_2} 
    + \underbrace{\eta^2_t\|f_t-\bar{f}_t\|^2}_{A_3} \\
    &=& \|\theta_t-\theta^*_t\|^2  -\underbrace{2\eta_t\langle\theta_t-\theta^*,\bar{f}_t\rangle}_{B_1} + \underbrace{\eta^2_t\|\bar{f}_t\|^2}_{B_2} \nonumber \\ & &  + A_2 + A_3.
\end{eqnarray}
Here, $\mathbb{E}[A_2] = 0$ due to $\mathbb{E}(f_t) = \bar{f}_t$, and $A_3$ is bounded according to Lemma~\ref{lem--1}. 
% Expanding $A_1$ we have
% %\begin{equation}\label{eq:lem1-1}
%     $A_1 = \|\theta_t-\theta^*_t\|^2-\underbrace{2\eta_t\langle\theta_t-\theta^*,\bar{f}_t\rangle}_{B_1} + \underbrace{\eta^2_t\|\bar{f}_t\|^2}_{B_2}$.
%\end{equation}
Note that $\bar{f}_t = \EB[f_t] = \EB[\nabla F(\theta_t)] = \nabla \EB[F(\theta_t)]$, and $\EB[F]$ inherits the $\mu$-strong convexity and L-smoothness from $F$.
By the L-smoothness of $\EB[F]$, we have
\begin{equation}
\|\bar{f}_t\|^2 \leq 2L(\EB[F(\theta_t)-{F}(\theta^*)]), 
\end{equation}
showing the boundness of $B_2$.
%\end{align}
%\noindent By convexity of $\|\cdot\|^2$ and due to L-smoothness of $F$, 
% which bounds 
%\begin{equation}
    % $B_2 \leq 2L\eta_t^2\EB[F(\theta_t)-F(\theta^*)]$.
%\end{equation}
Next, by the $\mu$-strong convexity of $\EB[F]$, we have
\begin{equation}
\langle \theta^* - \theta_t, \bar{f}_t \rangle \leq \EB[F(\theta^*)-F(\theta_t)] - \frac{\mu}{2}\|\theta_t-\theta^*\|^2, 
\end{equation}
proving the boundness of $B_1$.
% It follows that 
% $B_1 = -2\eta_t\langle\theta_t-\theta^*,\bar{f}_t\rangle \leq -2\eta_t\EB[F(\theta_t)-F(\theta^*)] -\mu\eta_t\|\theta_t-\theta^*\|^2$.
Applying the bounds of $B_1$ and $B_2$, we obtain \begin{eqnarray}
A_1 &\leq& (1- {\mu \eta_{t}})\|\theta_{t} -\theta^{*}\|^{2} \nonumber \\ & & -2\eta_{t}(1-L\eta_t)\EB[F(\theta_{t}) -F(\theta^{*})], 
\end{eqnarray}
where the last term on the RHS vanishes for $\eta_t < \frac{1}{L}$. Taking the expectation at both sides completes the proof.
% and since $0 \leq \eta_t \leq \frac{1}{L}$ bounds $C$ by zero,   
%     $A_1 \leq (1- \mu \eta_{t})\|\theta_{t} -\theta^{*}\|^{2}$.
% Finally we found the bounds of $A_1$ and $A_3$ to prove this Lemma. By taking the expectation to both sides of (\ref{eq16}), 
% \begin{equation}
%     \EB\|\theta_{t+1}-\theta^*\|^2\leq (1-\mu\eta_t)\EB\|\theta_t-\theta^*\|^2+ \eta^2_tB
% \end{equation}
% can be obtained.
% Finally, we have $\|{f_{t}} - \bar{f}_{t}\|^{2},$ unfolded as
% $\EB\|f_t-\bar{f}_t\|^2 \leq 4\sum^{2}_{i=1}w^2_{i}(\frac{1}{p_1}+\frac{1}{p_2})\frac{1}{K}\sum^K_{k=1}\sigma^2_k = 4\sum^{2}_{i=1}w^2_{i}(\frac{1}{p_1}+\frac{1}{p_2})\delta$.
\end{proof}
\subsection{Completing the Proof of Theorem 1}\label{sec:app_theorem}
\begin{proof}
Since $\eta_t = \frac{2}{\mu{t}+2L-\mu} \leq \frac{1}{L}$, applying Lemma~\ref{lem---2}, we have
\begin{equation}
    \Delta_{t+1} \leq \left(1 - \mu\eta_{t}\right)\Delta_{t} + \eta_{t}^{2}{B}.
\end{equation}

By induction, we aim to show that $\Delta_t \leq \frac{v}{t + 2\kappa-1}$ where $\kappa = \frac{L}{\mu}$ and $v = \max\{2\kappa\Delta_{1},{4B}/{\mu^2}\}$ as elaborated next.
By the definition of $v$, it is trivial that $\Delta_1\leq\frac{v}{2\kappa}$. Assuming that $\Delta_{t'} \leq \frac{v}{t' + 2\kappa-1}$ holds, we have
\begin{align}
    &\Delta_{t'+1} \leq (1-\mu\eta_{t'})\Delta_{t'} + \eta^{2}_{t'}B \\
    &\leq \left(1- \frac{2}{t'+2\kappa -1}\right) \frac{v}{t'+2\kappa-1}+\frac{{4B}/{\mu^2}}{(t'+2\kappa-1)^2}\\
    &= \frac{(t'+2\kappa-2)v-(v-{4B}/{\mu^{2}})}{(t'+2\kappa-1)^{2}}\\
    &\leq \frac{t'+2\kappa-2}{(t'+2\kappa-1)^{2}}v \leq \frac{v}{t'+2\kappa},
\end{align}
which proves that $\Delta_{t} \leq \frac{v}{t + 2\kappa-1}$. For $t=1$, we obtain
    \begin{equation}
    v = \max\{2\kappa\Delta_{1},\frac{4B}{\mu^2}\} \leq 2\kappa\Delta_{1} + \frac{4B}{\mu^2}.
    \end{equation}
    Finally, 
by the L-Smoothness of $F$, one has
    \begin{equation}
    \EB[F(\theta_{t})] - F^{*}  =\EB[F(\theta_{t}) -F(\theta^{*})] \leq \frac{L}{2}\EB\|\theta_{t} -\theta^{*}\|^{2}.
    \end{equation} 
    Applying Lemma~\ref{lem---2} with the results above, we have 
    \begin{equation}
    \EB\|\theta_{t} -\theta^{*}\|^{2} \leq \frac{v}{t +2\kappa -1} \leq \frac{2}{\mu}\cdot\frac{\mu L \Delta_1 + 2B}{\mu t+ 2L-\mu}, 
    \end{equation} 
    which completes the proof of the theorem.
\end{proof}

 \begin{IEEEbiographynophoto}{Won Joon Yun} is currently a Ph.D. student in electrical and computer engineering at Korea University, Seoul, Republic of Korea, since March 2021, where he received his B.S. in electrical engineering. He was a visiting researcher at Cipherome Inc., San Jose, CA, USA, during the summer of 2022; and also a visiting researcher at the University of Southern California, Los Angeles, CA, USA during the winter of 2022 for a joint project with Prof. Andreas F. Molisch at the Ming Hsieh Department of Electrical and Computer Engineering, USC Viterbi School of Engineering. 
% %\end{IEEEbiography}
 \end{IEEEbiographynophoto}
 \vspace{-5mm}

% %\begin{IEEEbiography}[{\includegraphics[width=1in,height=1.25in,clip]{people/people_yunseokkwak.png}}]{Yunseok Kwak} 
 \begin{IEEEbiographynophoto}{Yunseok Kwak}  
 was a Ph.D. student in electrical and computer engineering at Korea University, Seoul, Republic of Korea. He received his B.S. in mathematics from Yonsei University, Seoul, Republic of Korea, in 2021. % %\end{IEEEbiography}
 \end{IEEEbiographynophoto}

 \vspace{-5mm}

% %\begin{IEEEbiography}[{\includegraphics[width=1in,height=1.25in,clip]{people/people_hankyulbaek.png}}]{Hankyul Baek} 
 \begin{IEEEbiographynophoto}{Hankyul Baek} 
 is currently a Ph.D. student in electrical and computer engineering at Korea University, Seoul, Republic of Korea, since March 2021. He received his B.S. in electrical engineering from Korea University, Seoul, Republic of Korea, in 2020. He was with LG Electronics, Seoul, Republic of Korea, from 2020 to 2021. His current research interests include multi-agent deep reinforcement learning and its augmented reality applications. 
% %\end{IEEEbiography}
 \end{IEEEbiographynophoto}

 \vspace{-5mm}

% %\begin{IEEEbiography}[{\includegraphics[width=1in,height=1.25in,clip]{people/people_jungsoyi.png}}]{Soyi Jung} 
 \begin{IEEEbiographynophoto}{Soyi Jung} 
 (Member, IEEE) has been an assistant professor at the Department of Electrical of Computer Engineering, Ajou University, Suwon, Republic of Korea, since September 2022. Before joining Ajou University, she was an assistant professor at Hallym University, Chuncheon, Republic of  Korea, from 2021 to 2022; a visiting scholar at Donald Bren School of Information and Computer Sciences, University of California, Irvine, CA, USA, from 2021 to 2022; a research professor at Korea University, Seoul, Republic of Korea, in 2021; and a researcher at Korea Testing and Research (KTR) Institute, Gwacheon, Republic of Korea, from 2015 to 2016. She received her B.S., M.S., and Ph.D. degrees in electrical and computer engineering from Ajou University, Suwon, Republic of Korea, in 2013, 2015, and 2021, respectively. 
% Her current research interests include network optimization for autonomous vehicles communications, distributed system analysis, big-data processing platforms, and probabilistic access analysis. 

She was a recipient of Best Paper Award by KICS (2015), Young Women Researcher Award by WISET and KICS (2015), Bronze Paper Award from IEEE Seoul Section Student Paper Contest (2018), ICT Paper Contest Award by Electronic Times (2019), and IEEE ICOIN Best Paper Award (2021).
% %\end{IEEEbiography}
 \end{IEEEbiographynophoto}

 \vspace{-5mm}

% %\begin{IEEEbiography}[{\includegraphics[width=1in,height=1.25in,clip]{people/people_mingyueji.png}}]{Mingyue Ji} 
 \begin{IEEEbiographynophoto}{Mingyue Ji} 
 (Member, IEEE) received the B.E. degree in communication engineering from Beijing University of Posts and Telecommunications, China, in 2006, the M.Sc. degree in electrical engineering from the Royal Institute of Technology, Sweden, in 2008, the M.Sc. degree in electrical engineering from the University of California at Santa Cruz, in 2010, and the Ph.D. degree from the Ming Hsieh Department of Electrical Engineering, University of Southern California, in 2015. From 2015 to 2016, he was a Staff II System Design Scientist with Broadcom Corporation. He is an Assistant Professor with the Electrical and Computer Engineering Department\typeout{ and an Adjunct Assistant Professor with the School of Computing}, The University of Utah. 
 
 He received the IEEE Communications Society Leonard G. Abraham Prize for the Best IEEE Journal on Selected Areas in Communications Paper in 2019, the Best Paper Award in IEEE GLOBECOM 2021 Conference, the Best Paper Award in IEEE ICC 2015 Conference, the Best Student Paper Award in IEEE European Wireless 2010 Conference, and the USC Annenberg Fellowship from 2010 to 2014. Since 2020, he has been serving as an Associate Editor for IEEE Transactions on Communications. %His research interests include information theory, coding theory, the concentration of measure and statistics with the applications of caching networks, wireless communications, distributed storage and computing systems, distributed machine learning, and (statistical) signal processing.
% %\end{IEEEbiography}
 \end{IEEEbiographynophoto}

 \vspace{-5mm}

% %\begin{IEEEbiography}[{\includegraphics[width=1in,height=1.25in,clip]{people/people_mehdibennis.jpg}}]{Mehdi Bennis} 
 \begin{IEEEbiographynophoto}{Mehdi Bennis} 
 (Fellow, IEEE) is a tenured Full Professor with the Centre for Wireless Communications, University of Oulu, Finland, an Academy of Finland Research Fellow, and the Head of the Intelligent Connectivity and Networks/Systems Group (ICON). He has published more than 200 research papers in international conferences, journals, and book chapters. His main research interests are in radio resource management, heterogeneous networks, game theory, and distributed machine learning in 5G networks and beyond.

 He has been the recipient of several prestigious awards, including the 2015 Fred W. Ellersick Prize from the IEEE Communications Society, the 2016 Best Tutorial Prize from the IEEE Communications Society, the 2017 EURASIP Best Paper Award for the Journal of Wireless Communications and Networks, the All-University of Oulu Award for research, the 2019 IEEE ComSoc Radio Communications Committee Early Achievement Award, and the 2020 Clarivate Highly Cited Researcher by the Web of Science. He is an Editor of IEEE Transactions on Communications and the Specialty Chief Editor of Data Science for Communications in the Frontiers in Communications and Networks.
% %\end{IEEEbiography}
 \end{IEEEbiographynophoto}

 \vspace{-5mm}

% %\begin{IEEEbiography}[{\includegraphics[width=1in,height=1.25in,clip]{people/people_jihongpark.jpg}}]{Jihong Park} 
 \begin{IEEEbiographynophoto}{Jihong Park} 
 (Senior Member, IEEE) received the B.S. and Ph.D. degrees from Yonsei University, South Korea. He is currently a Lecturer (Assistant Professor) with the School of Information Theory, Deakin University, Australia. 
 His research interests include ultra-dense/ultra-reliable/mmWave system designs, and distributed learning/control/ledger technologies and their applications for beyond-5G/6G communication systems. He served as a Conference/Workshop Program Committee Member for IEEE GLOBECOM, ICC, and WCNC, and for NeurIPS, ICML, and IJCAI. He is an Associate Editor of Frontiers in Data Science for Communications, and a Review Editor of Frontiers in Aerial and Space Networks.
% %\end{IEEEbiography}
 \end{IEEEbiographynophoto}

 \vspace{-5mm}
% %\begin{IEEEbiography}[{\includegraphics[width=1in,height=1.25in,clip]{people/joongheonkim.eps}}]{Joongheon Kim}
 \begin{IEEEbiographynophoto}{Joongheon Kim} 
 (Senior Member, IEEE) has been with Korea University, Seoul, Korea, since 2019, and he is currently an associate professor. He received the B.S. and M.S. degrees in computer science and engineering from Korea University, Seoul, Korea, in 2004 and 2006, respectively; and the Ph.D. degree in computer science from the University of Southern California (USC), Los Angeles, CA, USA, in 2014. Before joining Korea University, he was with LG Electronics (Seoul, Korea, 2006--2009), InterDigital (San Diego, CA, USA, 2012), Intel Corporation (Santa Clara in Silicon Valley, CA, USA, 2013--2016), and Chung-Ang University (Seoul, Korea, 2016--2019). 

 He serves as an editor for \textsc{IEEE Transactions on Vehicular Technology}, \textsc{IEEE Transactions on Machine Learning in Communications and Networking}, \textsc{IEEE Communications Standards Magazine}, \textit{Computer Networks (Elsevier)}, and \textit{ICT Express (Elsevier)}. He is also a distinguished lecturer for \textit{IEEE Communications Society (ComSoc)} (2022-2023) and \textit{IEEE Systems Council} (2022-2024).

 He was a recipient of Annenberg Graduate Fellowship with his Ph.D. admission from USC (2009), Intel Corporation Next Generation and Standards (NGS) Division Recognition Award (2015), \textsc{IEEE Systems Journal} Best Paper Award (2020), IEEE ComSoc Multimedia Communications Technical Committee (MMTC) Outstanding Young Researcher Award (2020), IEEE ComSoc MMTC Best Journal Paper Award (2021), and Best Special Issue Guest Editor Award by \textit{ICT Express (Elsevier)} (2022). He also received numerous awards from IEEE international conferences including IEEE ICOIN Best Paper Award (2021), IEEE Vehicular Technology Society (VTS) Seoul Chapter Awards for APWCS (2019 and 2021), and IEEE ICTC Best Paper Award (2022). 
% %\end{IEEEbiography}
 \end{IEEEbiographynophoto}

\end{document}